\documentclass[preprint,review,10pt]{elsarticle}

\usepackage{amssymb,geometry,xcolor,amsmath,bbm,multirow,bbold,mathtools,amssymb,graphics,graphicx,epstopdf,ntheorem,diagbox,enumitem,diagbox}
\usepackage{xr}
\usepackage{booktabs} 
\newtheorem{lemma}{Lemma}
\newtheorem{theorem}{Theorem}
\newtheorem{corollary}{Corollary}
\newtheorem{proposition}{Proposition}
\newtheorem*{proof}{Proof}

\usepackage{color}
\newcommand{\zcolor}[1]{{\color{black}#1}}
\newcommand{\xcolor}[1]{{\color{black}#1}}

\journal{Pattern Recognition}

\begin{document}

\begin{frontmatter}



\title{\zcolor{Subspace-Constrained
Quadratic Matrix Factorization: Algorithm and Applications}}


\author[bnu]{Zheng Zhai}
\author[HH]{Xiaohui Li\corref{hh}}
\affiliation[bnu]{organization={Department of Statistics, Faculty of Arts and Sciences, Beijing Normal University},
addressline={No.18 Jinfeng Road}, city={Zhuhai},postcode={519087}, state={Guangdong},country={China}}
\affiliation[HH]{organization={School of Mathematics and Information Sciences, Yantai University},
addressline={No.30 Laishan Qingquan Road}, 
city={Yantai},
postcode={264005}, 
state={Shandong},
country={China}}
\cortext[hh]{Corresponding author: Xiaohui Li (lxhmath@zju.edu.cn).}

\begin{abstract}
\zcolor{
Matrix Factorization has emerged as a widely adopted framework for modeling data exhibiting low-rank structures. To address challenges in manifold learning, this paper presents a subspace-constrained quadratic matrix factorization model. The model is designed to jointly learn key low-dimensional structures, including the tangent space, the normal subspace, and the quadratic form that links the tangent space to a low-dimensional representation. We solve the proposed factorization model using an alternating minimization method, involving an in-depth investigation of nonlinear regression and projection subproblems. Theoretical properties of the quadratic projection problem and convergence characteristics of the alternating strategy are also investigated.
To validate our approach, we conduct numerical experiments on synthetic and real-world datasets. Results demonstrate that our model outperforms existing methods, highlighting its robustness and efficacy in capturing core low-dimensional structures.}
\end{abstract}


\bigskip


\end{frontmatter}



\section{Introduction}
Matrix factorization has achieved remarkable success across various domains, including clustering~\citep{clust}, recommendation systems~\citep{recom}, graph learning~\citep{graph}, and factor analysis~\citep{facto}. The classical linear matrix factorization models~\citep{mf1} can be expressed as \zcolor{$X \approx AB$, where $X=[x_1,x_2,...,x_n]$ stacks the samples by columns, the columns of $A$ represent the basis, and each column of $B$ represents the coordinates under the basis $A$. }
However, this model has its limitations, as it can only discern the linear structures within $X$. In cases where the data showcases nonlinear, low-dimensional properties, the linear representation model will fall short of capturing more intrinsic details~\citep{ltsa}. In specific, we assume
\begin{equation}\label{assu}
    x_i=f(\tau_i)+\epsilon_i, \ i = 1,...,n,
\end{equation}
where $f(\cdot):{\mathbb R}^d\rightarrow {\mathbb R}^D$ is a nonlinear transformation from the low-dimensional representation to the ambient high-dimensional space and $\epsilon_i$ is the noise which obeys some distribution. To reduce the dimension of searching space for $f(\cdot)$, we restrict $f(\cdot)$ to the subspace-constrained quadratic function class.
The subspace-constrained quadratic function class can help us extract latent structural insights, including the origin, tangent space, normal space, and even the second fundamental form, through an optimization model. The rationale behind this model is that, regardless of the complexity of the distribution of $\{x_i\}$, we can at least select a subset of $\{x_i\}$ in the local neighborhood \zcolor{that exhibits quadratic characteristics.}
In other words, we learn $f(\cdot)$ from the subspace-constrained quadratic function class of
\[
{\cal F} = \{f(\tau) = c+U\tau+V{\cal A}(\tau,\tau), U^TU = I_d, V^T V=I_s, U^TV = \bf 0 \},
\]
where the columns of $U\in {\mathbb R}^{D\times d}$ consist of the basis on the tangent space, \xcolor{and} the columns of $V\in {\mathbb R}^{D\times s}$ span a $s$-dimensional subspace in the normal space. Here, the tensor ${\cal A} \in {\mathbb R}^{s}\times {\mathbb R}^d\times {\mathbb R}^d$ can be viewed as a \zcolor{bilinear} operator ${\cal A}(\cdot,\cdot):{\mathbb R}^d\times {\mathbb R}^d\rightarrow {\mathbb R}^{s}$.  
We set $s$ to be less than  $(d^2+d)/2$. This choice is motivated by the fact that the quadratic form of $\tau\in {\mathbb R}^d$ produces results that are linear combinations of $(d^2+d)/2$ cross terms in the form of $\{\tau_i\tau_j, i,j=1:d\}$. Meanwhile, since the dimension of the normal space is $D-d$, it also necessitates that $s \leq D-d$.


\zcolor{To the best of our knowledge, few studies have delved into estimating the subspace information and the second-order information within the distribution of the data.
However, most of these studies, such as ~\citep{Vincent} and~\citep{Aamari}, primarily focus on theoretically estimating the tangent space or the second fundamental form from statistical perspectives, leaving computational challenges unresolved.} We propose a novel approach to simultaneously learn the tangent space, a subspace of the normal space, the second fundamental form, and the low-dimensional representation ${\tau_i}$ via the optimization problem:
\begin{equation}\label{approx}
\min_{f\in {\cal F},\{\tau_i\in {\mathbb R}^d\}} \sum_{i=1}^n \|x_i-f(\tau_i)\|_2^2.
\end{equation}
This model can be further reformulated as a special matrix factorization problem. Recalling the definition of $\cal F$ and the orthogonal relationship between $U$ and $V$, we can rewrite \eqref{approx} as:
\begin{equation}\label{mini_prob}
     \min_{c, {\cal A},\{\tau_i\}\atop U^T U = I_d} \min_{V\in {\mathbb F}_{D,s}(U)}\sum_{i=1}^n \|x_i - (c+U\tau_i+V {\cal A}(\tau_i,\tau_i))\|_2^2,
\end{equation}
where ${\mathbb F}_{D,s}(U)=\{V|V\in {\mathbb R}^{D\times s}, V^TV=I_{s}, V^TU = \bf 0\}$.
In fact, the \zcolor{two-phase} optimization problem can be further simplified into a \zcolor{one-phase} optimization by introducing an orthonormal matrix $Q:=[U,V] \in {\mathbb R}^{D\times (d+s)}$ such that $Q^TQ = I_{d+s}$. Additionally, we simplify the quadratic form by introducing a matrix $\Phi = [\tau_1,...,\tau_n]\in \mathbb{R}^{d\times n}$, a function $\Psi(\cdot):\mathbb{R}^{d\times n}\rightarrow \mathbb{R}^{\frac{d^2+d}{2}\times n}$ defined such that each column \zcolor{consists} of the cross-product of each coordinate of $\tau_i$, i.e., ${\Psi(\Phi)}_{\cdot,i} = [\tau_{i,1}^2,\tau_{i,1}\tau_{i,2},...,\tau_{i,d}^2]^T$, and a matrix $\Theta\in \mathbb{R}^{\frac{d^2+d}{2}\times s}$ such that $\Theta^T\Psi(\Phi) = [{\cal A}(\tau_1,\tau_1),...,{\cal A}(\tau_n,\tau_n)]^T$. Then, the objective function in \eqref{mini_prob} can be reformulated as:
\begin{equation}\label{factor_model}
\begin{aligned}
\min_{\Theta,c, \Phi,Q^TQ = I_{d+s}} \ell(\Theta, c, Q,\Phi) 
=\|X- [c, Q]
\left[
{\bf 1}_n, \Phi^T, \Psi^T(\Phi) \Theta
\right]^T
\|_{\rm F}^2.
\end{aligned}
\end{equation}
The optimization in \eqref{factor_model} can be viewed as a generalized matrix factorization problem, where $X$ is factorized into $RF_{\Phi,\Theta}$, with $R:=[c,Q]$ and $F_{\Phi,\Theta}:=\left[ {\bf 1}_n, \Phi^T, \Psi^T(\Phi) \Theta \right]^T$. This factorization is subject to the subspace constraint $Q^TQ = I_{d+s}$. Hence, we refer to \eqref{factor_model} as the subspace-constrained quadratic matrix factorization problem (SQMF).

As $\Psi(\Phi)$ is a higher-order term of $\Phi$, we aim to regulate the scale of $V\Theta^T \Psi(\Phi)$ such that it acts as a modification of the linear term $U\Phi$. Therefore, we introduce a regularization variant of \eqref{factor_model} to strike a balance between the higher-order fitting error and the contribution of the second-order term.
\begin{equation}\label{reg}
\min_{\Theta,c, \Phi,Q^TQ = I_{d+s},} \ell_{\lambda}(\Theta, c, Q,\Phi) 
=\|X- [c, Q]
\left[
{\bf 1}_n, \Phi^T, \Psi^T(\Phi) \Theta
\right]^T
\|_{\rm F}^2 + \lambda \|\Theta^T \Psi(\Phi)\|_{\rm F}^2.
\end{equation}

We term the model in \eqref{reg} the regularized subspace-constrained quadratic matrix factorization (RSQMF) problem. The regularization imposed on the second-order term empowers us with manual control over the curvature of the quadratic function. Consequently, it plays a pivotal role in mitigating the overfitting phenomenon and enhances the stability of the optimization process.

Our contributions are fourfold. First, we define the subspace-constrained quadratic matrix factorization (SQMF) problem and introduce an alternating optimization algorithm to solve it. Second, we extend this formulation by proposing the regularized variant, RSQMF, which incorporates regularization to enhance model performance. Third, we establish key theoretical properties for solving SQMF, providing insights into the optimality and behavior of the solution. Finally, we demonstrate the versatility of RSQMF by applying it to various tasks, including tangent space estimation, data refinement and reconstruction, and learning latent representations.

\begin{figure}[t] 
\centering
\includegraphics[width=\linewidth]{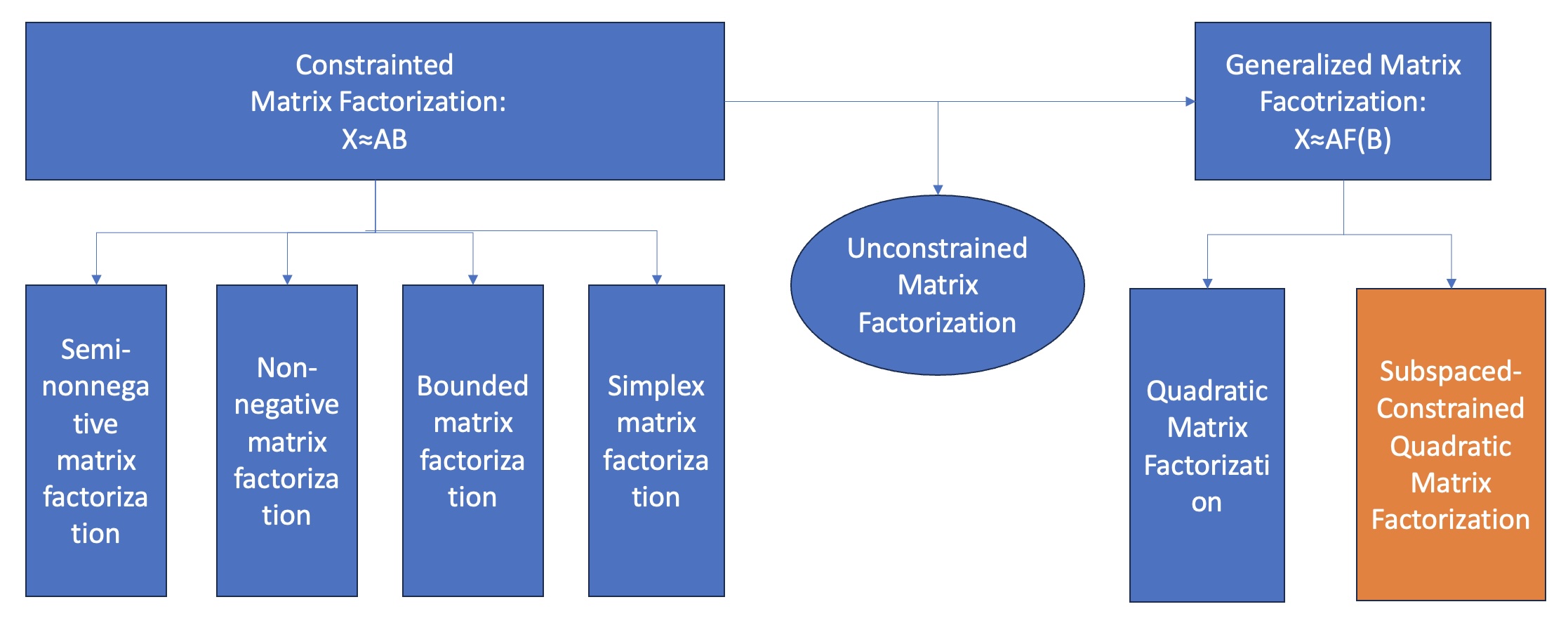} 
\vspace{-12mm}
\caption{Illustration of the matrix factorization landscape and the relationship between SQMF and prior works. \label{fig:MF_relationship}}
\end{figure}

\subsection{Related works}
We utilize subspace-constrained quadratic matrix factorization to extract and understand the manifold structure underlying the data. To effectively demonstrate the significance of our approach, it is crucial to provide an overview of the key methods within both matrix factorization and manifold learning. This will help position our work within the broader context, illustrating how our method integrates principles from both domains to capture the complex geometric relationships inherent in the data.
\subsubsection{Landscape for Matrix Factorization}
We categorize the global matrix factorization landscape into three distinct types: constrained matrix factorization, unconstrained matrix factorization, and generalized matrix factorization, as illustrated in Figure \ref{fig:MF_relationship}.

\textbf{Unconstrained Matrix Factorization:} 
The unconstrained factorization model, as presented in~\citep{MF}, seeks to optimize the following objective: $\min_{A \in \mathbb{R}^{r \times s}, B \in \mathbb{R}^{s \times c}} \|X - AB\|_{\rm F}^2$. This model can be efficiently solved using Singular Value Decomposition (SVD) of $X$, or through the eigenvalue decomposition of $XX^T$ and $X^TX$. The absence of constraints provides greater flexibility, allowing the model to minimize factorization loss by freely adjusting both $A$ and $B$. 

\textbf{Constrained Matrix Factorization:} Introducing constraints on  $A$  or  $B$  limits the solution space, which typically leads to an increase in factorization loss due to the reduced flexibility. Various forms of constrained matrix factorization include nonnegative matrix factorization (NMF)~\citep{mf2}, semi-nonnegative matrix factorization (Semi-NMF)~\citep{smf}, simplex matrix factorization (SMF)~\citep{simplex}, bounded simplex matrix factorization (BSMF)~\citep{Boundedsimplex} and \zcolor{ manifold regularized matrix factorization (MMF)~\citep{ZhangZhao}. These models yield solutions tailored to different objectives: for instance, MMF produces a solution that preserves connectivity, while NMF generates one with improved clustering properties.}

\textbf{Generalized Matrix Factorization:} Generalized matrix factorization extends beyond linear terms in  $B$  by incorporating nonlinear functions ${F}(B)$, such as quadratic constructions. 
This advanced approach enhances the model’s ability to capture deeper, more intrinsic features within the data, resulting in a more accurate fit compared to traditional unconstrained models.
An example of this is the quadratic matrix factorization model, as discussed in~\citep{Zhai}.
\subsubsection{Manifold Fitting/Learning}
Since we use our subspaced-constrained quadratic matrix factorization to learn the latent manifold, we also provide a concise overview of manifold fitting or denoising methods, encompassing various approaches such as local linear fitting (LPCA)~\citep{lpca}, ridge-related techniques like KDE~\citep{kde} and LOG-KDE~\citep{msf}, tangent space estimation through weighted summation (MFIT)~\citep{mfit}, and higher-order regression methods like moving least squares (MLS)~\cite{mls}. For the sake of comparison, we delve into \zcolor{seven} methods in detail.

The most straightforward model for understanding the latent linear structure among \zcolor{datasets} is local PCA (LPCA)~\citep{lpca} which fits the data with a model of $f(\tau)= c+ U\tau$ where $c$ is the center and $U$ is orthonormal matrix. The estimator can be acquired through two steps: First, learning the center, such as $c_x = \frac{1}{m}\sum_{k=1}^m  x_{i_k}$, where $\{x_{i_k}\}$ are the samples in the local neighborhood. Second, computing the shifted covariance matrix $M = \frac{1}{m}\sum_{k = 1}^m (x_{i_k} -c_x)(x_{i_k}-c_x)^T$. \zcolor{Let $P$ denote the projection matrix associated with the subspace spanned by the $d$ principal eigenvectors of $M$. The denoised version of $x$, representing a point on the estimated manifold projected from $x$, is given by $x_{\textnormal{new}}=c_x+P(x-c_x)$.}

Apart from the local linear fitting methods, we can denoise or refine the data next to manifold by defining a subset that adheres to some conditions such as the ridge defined through kernel density estimation (KDE)~\citep{kde} or logarithm kernel density estimation (LOG-KDE)~\citep{msf}. Specifically,
the ridge~\citep{kde} is delineated via the Hessian and gradient of the density function $p(x)$ as 
\begin{equation}\label{equ:ridge}
\begin{aligned}
{\cal R}(p)\coloneqq \{x\mid &\Pi^{\perp}(\nabla^2{p}(x)) \nabla {p}(x) = {\bf 0}, \lambda_{d+1}(\nabla^2 {p}(x))<0\},
\end{aligned}
\end{equation}
\zcolor{where  $p(x)$  can be constructed nonparametrically. For example,  $p(x) = \frac{1}{n} \sum_{i=1}^n K_h(x, x_i)$, where  $K_h(\cdot, \cdot)$  is a kernel function with bandwidth  $h$. Additionally,  $\Pi^{\perp}(\nabla^2 \hat{p}(x)) = I - UU^T$ , where the columns of  $U$  are the leading  $d$  eigenvectors of  $\nabla^2 \hat{p}(x)$.}
The ridge \eqref{equ:ridge} typically lacks closed-form solutions. In such cases, the subspace-constrained mean shift (SCMS) algorithm~\citep{msf} is commonly employed to iteratively identify the ridge set.
The LOG-KDE~\citep{msf} ridge estimation method is essentially a \zcolor{kernel-based} ridge estimation approach with $p(x)$ substituted by $\log p(x)$. This modification impacts the covariance matrix and consequently results in a different estimation of the tangent space.

\zcolor{Similar to the ridge approach, alternative methods can be employed to define a smooth manifold under specific conditions related to the projection matrix. One notable method is manifold fitting (MFIT)~\citep{mfit}, which constructs a manifold from noisy data using the following formulation:}
$
\{x\mid\Pi \big(\sum_{i}\alpha(x,x_i)\Pi_i(x-x_i)\big)={\bf 0}\},
$
where $\alpha(x,x_i)$ represents a weight function, $\Pi_i$ denotes the projection matrix related with the normal space at $x_i$, and $\Pi$ stands for the projection matrix corresponding to space spanned by the leading $D-d$ eigenvectors of $\sum_{i}\alpha(x,x_i)\Pi_i$. The function $\alpha(x,x_i)$ can be selected as $\frac{1}{\kappa}(1-\frac{\|x-x_i\|_2^2}{r^2})^\gamma$, where $\kappa$ is a normalized factor ensuring $\sum_i \alpha(x,x_i)=1$, $r$ represents the radius, and $\gamma$ is the power parameter. 

Additionally, we would like to discuss a relevant methodology for higher-order manifold estimation: the moving least squares method (MLS)~\cite{mls}. This approach encompasses two key stages: identifying the appropriate tangent space and executing the regression step. Specifically, for any $x\in\mathbb{R}^D$, it initially seeks the $d$-dimensional hyperplane 
$\cal H$ in $\mathbb{R}^{D}$
that minimizes the following expression:
\[
{\cal L}_1({\cal H})=\min_{q\in{\cal H},x-q\perp {\cal H}}\sum_{i} \alpha(q, x_i) \rho^2(x_i, {\cal H}),
\]
where $\alpha(\cdot,\cdot)$ is a weight function and $\rho(x_i,{\cal H})$ is the  distance between $x_i$ and  ${\cal H}$. Subsequently, a coordinate system is established on $\cal H$ with the origin $q$ being the projection point of $x$ onto $\cal H$.
Utilizing this coordinate system, we can obtain the $d$-dimensional representation  $\tau_i$ of $x_i$ by projecting $x_i$ to $\cal H$. In the second step, a polynomial function $p(\tau):\mathbb{R}^d\to\mathbb{R}^D$ of a specified degree is fitted by minimizing the following weighted squares: ${\cal L}_2(p)=\sum_i\alpha(q,x_i)\|p(\tau_i)-x_i\|_2^2$.
The origin point of $p(\tau)$ at $\tau = \bf 0$ is considered as the refined or denoised point of $x$.

Spherical PCA (SPCA)~\citep{Spherelets}  is a variant of LPCA which first learns a $d+1$ dimensional affine space parameterized by $V$ and \zcolor{projects} the data into the $d+1$ affine space via $y_i = {\bar x}+VV^T (x_i-{\bar x})$.
Then, the method learns a $d$-dimensional sphere parameterized by the center $c$ and radius $r$ can be learned through:
\[
(\hat{c},\hat{r}) = \arg\min_{c,r} \sum_{i=1}^n ((y_i-c)^T(y_i-c)-r^2).
\]
After we obtain $\hat{c},\hat{r}$, for any data $x$ outside the sphere and $VV^T(x-c)\neq \bf 0$, the projected point on the sphere can be achieved by:
$
P(x) = \hat{c} + \frac{\hat{r}}{\|VV^T(x-\hat{c})\|}{VV^T(x-\hat{c})}
$.

Structure-Adaptive Manifold Estimation (SAME)~\citep{SAME} is a diffusion method which aims to recover data $\{x_1,...,x_n\}$ from noisy observations $\{y_1,...,y_n\}$ using  weighted averaging. The weights at $x_i$ are computed based on a suitable projector $\Pi_i$. Denote by $\Pi_i^t$ the estimation of $\Pi_i$ in the $t$-th iteration.
Then the update of $x_i$ at the $t$-th iteration is given by
\[
x^t_i = \frac{\sum_j w^t_{i,j}y_j}{\sum_j w^t_{i,j}},
\]
where weights are given by 
$
w_{i,j}^t = K(\frac{\|\Pi_i^t(y_i-y_j)\|^2}{h_t^2})I(\|y_i-y_j\|\leq \tau),
$ where 
$h_t=a^{-t}h_0$ is the $t$-th bandwidth with a rate $a>1$, $h_0$ is the initial bandwidth, and $\tau>0$ is a threshold. In SAME, $\Pi^t$ is updated using the eigenvalue decomposition of the covariance matrix 
$\Sigma_i^t=\sum_{j\in J_i^t} (x_j^t-x_i^t)(x_j^t-x_i^t)^T$, where  $J^t_i=\{j\mid \|x^k_j-x^k_i\|\leq \gamma/a^k\}$ is an index set with $\gamma>0$.

Finally, it is also worth noting that the unconstrained regularized quadratic matrix factorization (RQMF) model has been explored in our prior research~\citep{Zhai}, wherein we propose a model represented as $f(\tau) = c + A\tau + \mathcal{B}(\tau,\tau)$, with $A \in \mathbb{R}^{D\times d}$ and $\mathcal{B} \in \mathbb{R}^{D\times d\times d}$. However, the proposed model in prior work diverges from the current model in two significant respects:
Firstly, unlike the model proposed in~\citep{Zhai}, where $A$ and $\mathcal{B}$ are treated as unconstrained parameters, we introduce explicit orthonormal restrictions on $U$ and $V$. This imposition helps mitigate the correlation between the linear and quadratic terms, thereby enhancing the model's fitting performance and mitigating the risk of over-fitting.
Secondly, we reduce the \zcolor{number} of free parameters by substituting the unconstrained tensor $\mathcal{B}$ with an orthonormal matrix $V \in \mathbb{R}^{D\times s}$ and an unconstrained tensor $\mathcal{A} \in \mathbb{R}^{s\times d\times d}$. This substitution notably diminishes the parameter search space, especially when $s$ is much smaller than $D$, resulting in a more computationally efficient model. These modifications offer several advantages, including the development of more robust algorithms for optimizing the objective function.
\subsection{Paper Organization}
The rest of the paper proceeds as follows: In Section 2, we propose an alternating algorithm to solve the factorization model presented in \eqref{factor_model}. Specifically, we partition the factorization model into a regression sub-problem concerning $\Theta, c,$ and $Q$, and a projection sub-problem concerning $\Phi$. In Section 3, we study the convergence properties of the alternating minimization algorithm for solving the regression (inner iteration) and fitting model (outer iteration). In Section \ref{application}, we elaborate on three applications for the SQMF model. Subsequently, in Section \ref{experiment}, we present numerical simulations on both synthetic and real datasets, demonstrating the superiority of the SQMF over related methods. 
\subsection{Notation}
\begin{table}[t]
\centering
\caption{List of Symbols \label{tab:section1-symbols}}
\resizebox{\linewidth}{!}{
\begin{tabular}{c|c} \toprule 
 \textbf{Symbol} & \textbf{Description} \\ \hline
${\cal A}$ & Three-order tensor which represents the second fundamental form, which is also \zcolor{an} operator ${\mathbb R}^{d\times d}\rightarrow {\mathbb R}^s$.  \\
$\Theta$ & Matrix with the shape of ${(d^2+d)}/{2}\times s$ , which is the matrix form of $\cal A$.\\
$U,V,Q$ & $U$ represents the basis for the tangent space and $V$ represents the basis for the normal space. $Q:=[U,V]$.\\
$c$ &  Shifting center for the coordinate system. \\
$\Phi$ & The coordinate matrix whose $i$-th column represents the coordinate for $x_i$. \\
$(Q^*, c^*, \Theta^*)$ & Accumulation point of the sequences $\{ Q_t,c_t,\Theta_t\}$ \\
$\Psi(\Phi)$ & The quadratic coordinate matrix, whose $i$-th column consists of the cross-product from the entries of $\tau_i$.\\ \bottomrule
\end{tabular}}
\end{table}

{We present a concise overview of \zcolor{the} frequently used notations in Table \ref{tab:section1-symbols}.
The three-order tensor $\cal A$ with the shape of ${\mathbb R}^{s}\times \mathbb R^d\times \mathbb R^d$ is symmetric for its second and third \zcolor{dimensions}, i.e., each of its \zcolor{slices} $\{A_k:={\cal A}(k,:,:),k = 1,...,s\}$ is a symmetric matrix in ${\mathbb R}^d\times {\mathbb R}^d$. ${\cal A}$ can also be seen as an operator from ${\mathbb R}^d\times{\mathbb R}^d$ to ${\mathbb R}^s$ such that the $k$-th element of ${\cal A}(\tau,\tau)$ is $\tau^T {A}_k \tau$. 
The matrix function $\Psi(\Phi): {\mathbb R}^{d\times n}\rightarrow {\mathbb R}^{\frac{d^2+d}{2}\times n}$ maps the $i$-th column of $\Phi$ to the $i$-th \zcolor{column} of $\Psi(\Phi)$ such as $\{\Psi(\Phi)\}_{\cdot,i} = [\tau_{i,1}^2,\tau_{i,1}\tau_{i,2},...,\tau_{i,d}^2]^T$. The adjoint operator of $\mathcal{A}$, denoted as ${\cal A}^*(c):{\mathbb R}^{s}\rightarrow {\mathbb R}^{d\times d}$, is defined as ${\mathcal{A}}^*(c) = \sum_{k=1}^{s} c_k A_k,  \forall, c\in {\mathbb R}^{s}$, where $A_k$ represents the $k$-th slice of $\cal A$. Let $A_\tau$ \zcolor{represent} the action of $\cal A$ on $\tau$, \zcolor{defined as} $A_\tau = [A_1\tau, A_2\tau,...,A_s\tau]^T\in {\mathbb R}^{s\times d}$.
We denote the operator norm and $\ell_2$ norm as $\|\cdot\|_{\text{op}}$ and $\|\cdot\|_2$ respectively. Additionally, $\sigma_k(A)$ denotes the $k$-th largest singular value of matrix $A$.}

\section{Subspace-Constrained Quadratic Matrix Factorization}
In this section, we begin by establishing the loss bound for the RSQMF model. Following that, we introduce the algorithm for solving both SQMF and RSQMF models.
\subsection{Solution Properties}
Here, we study the relationship between three matrix factorization models: the unconstrained matrix factorization (MF), the subspace-constrained quadratic matrix factorization (SQMF) and the regularized subspace-constrained quadratic matrix factorization (RSQMF). \zcolor{For ease of notation, we introduce two sets:} $\Omega = \{(\Theta,c,Q,\Phi)|Q^TQ = I_{d+s}, {\Phi}\in {\mathbb R}^{d\times n},{c}\in {\mathbb R}^D,\Theta = {\bf 0}\in {\mathbb R}^{\frac{d^2+d}{2}\times s}\}$ and ${\cal D} = \{(\Theta,c,Q,\Phi)|Q^TQ = I_{d+s}\} $.
\begin{proposition}   
\label{bound_relationship}
Let $(\widehat{Q},\widehat{c},\widehat{Q},\widehat{\Phi})$, $(\Theta_\lambda,c_\lambda,Q_\lambda,\Phi_\lambda)$ and $({\Theta'},{c'},{Q'},{\Phi'})$ be the optimal solution of RSQMF with extra constraint $\Omega$,
the RSQMF model and the SQMF model, respectively.  
Then, we have:
\[
\ell(\Theta',c',Q',\Phi')\leq \ell(\Theta_\lambda,c_\lambda,Q_\lambda,\Phi_\lambda)\leq\ell(\widehat{\Theta},\widehat{c},\widehat{Q},\widehat{\Phi}).
\]
\begin{proof}
    The first inequality is straightforward because $(\Theta',c',Q',\Phi')$ is the optimal solution of  SQMF. \zcolor{Next, we derive the second inequality.}
    Recall that $(\Theta_\lambda,c_\lambda,Q_\lambda,\Phi_\lambda)$ minimizes $\min_{(\Theta,c,Q,\Phi)\in {\cal D}} \ell_\lambda(\Theta,c,Q,\Phi)$, and $(\widehat{Q},\widehat{c},\widehat{Q},\widehat{\Phi})$ minimizes $\min_{(Q,c,\Theta,\Phi)\in\Omega} \ell_\lambda(\Theta,c,Q,\Phi)$. Minimizing the same objective under the restriction of $\Omega$ leads to an increase in the objective function:
    \[  \ell_\lambda(\Theta_\lambda,c_\lambda,Q_\lambda,\Phi_\lambda) \leq \ell_\lambda(\widehat{\Theta},\widehat{\lambda},\widehat{Q},\widehat{\Phi}),
    \]
    which is equivalent to:
\[
\|X- [c_\lambda, Q_\lambda]
\left[
{\bf 1}_n, \Phi_\lambda^T, \Psi^T(\Phi_\lambda) \Theta_\lambda
\right]^T
\|_{\rm F}^2 + \lambda \|\Theta_\lambda^T \Psi(\Phi_\lambda)\|_{\rm F}^2\leq \|X- [\widehat{c}, \widehat{Q}]
\left[
{\bf 1}_n, \widehat{\Phi}^T, \Psi^T(\widehat{\Phi}) \widehat{\Theta}
\right]^T
\|_{\rm F}^2.
\]
By eliminating the nonnegative item $\lambda \|\Theta_\lambda^T \Psi(\Phi_\lambda)\|_{\rm F}^2$, we obtain the second inequality.
\end{proof}
\end{proposition} 

\zcolor{Notably, when the RSQMF model is restricted to  $\Omega$, it simplifies to an unconstrained matrix factorization model.} The proposition \ref{bound_relationship} indicates that although the regularized version of RSQMF increases the fitting error compared with SQMF, the error for RSQMF is still smaller than that of the unconstrained matrix factorization model, regardless of the choice of $\lambda$ ($\lambda>0$).


\subsection{Algorithm}
This section explores the application of \zcolor{the} alternating minimization algorithm to address SQMF in \eqref{factor_model} and RSQMF in \eqref{reg}. As SQMF can be viewed as a special form of RSQMF with the tunable parameter $\lambda=0$, we just need to discuss the differences between SQMF and RSQMF when necessary.

To comprehensively tackle SQMF and RSQMF, we introduce two pivotal subproblems. The first subproblem is formulated by fixing $\Phi$ at $\Phi^{k-1}$ and optimizing the remaining parameters within $\ell(\Theta, c, Q, \Phi^{k-1})$ or $\ell_\lambda(\Theta, c, Q, \Phi^{k-1})$. Subsequently, the second subproblem is defined by fixing $(\Theta, c, Q)$ as $(\Theta^k, c^k, Q^k)$ and optimizing $\ell(\Theta^k, c^k, Q^k, \Phi)$ or $\ell_{\lambda}(\Theta^k, c^k, Q^k, \Phi)$ with respect to $\Phi$. In other words, we iteratively solve the following two subproblems:

\begin{equation}\label{step12}
\begin{aligned}
(\Theta^k,c^k,Q^k) = & \arg\min_{\Theta,c, Q^TQ = I_{d+s}} \ell(\Theta, c, Q, \Phi^{k-1}),\\
\Phi^k = &  \arg\min_{\Phi} \ell(\Theta^k, c^k, Q^k, \Phi).
\end{aligned}
\end{equation}
If the sequence $\{\Theta^k, c^k, Q^k, \Phi^k\}_{k=1}^\infty$ converges, we designate the convergence point as the solution to the joint minimization problem \eqref{factor_model} and utilize the convergence point $\{\Theta^*, c^*, Q^*, \Phi^*\}$ as the optimal solution. Since we also employ an alternating strategy to solve each of the subproblems in \eqref{step12}, we refer to the iteration in \eqref{step12} as an outer iteration. Next, we will analyze and optimize both $\min_{\Theta,c, Q^TQ = I_{d+s}} \ell(\Theta, c, Q, \Phi^{k})$ and $\min_{\Phi} \ell(\Theta^k, c^k, Q^k, \Phi)$ in subsections \ref{sub1} and \ref{sub2} respectively. 
From the iterations in \eqref{step12}, it is evident that the function value decreases with the updating of the sequence $\{\Theta^k, c^k, Q^k, \Phi^k\}$, i.e.,
\begin{equation}\label{dec}
\ell(\Theta^k, c^k, Q^k, \Phi^{k}) \leq \ell(\Theta^k, c^k, Q^k, \Phi^{k-1}) \leq \ell(\Theta^{k-1}, c^{k-1}, Q^{k-1}, \Phi^{k-1}).
\end{equation}
Recalling that $\ell(\Theta^k, c^k, Q^k, \Phi^{k}) \geq 0$ for all $k$, we can conclude that the sequence $\{\ell(\Theta^k, c^k, Q^k, \Phi^{k})\}_{i=1}^{\infty}$ converges by the monotone convergence theorem. The decreasing relationship expressed in \eqref{dec} further indicates that the model \eqref{factor_model} improves in precision with \zcolor{each successive iteration step $k$}.

\subsection{Alternating Method for Minimizing $\ell(\Theta, c, Q, \Phi^{k})$\label{sub1}}
In this subsection, we fix $\Phi$ as $\Phi^k$ and give an alternating approach to solve the first subproblem in \eqref{step12}. First, we can reformulate the problem $\min_{\Theta,c, Q^TQ = I_{d+s}} \ell(\Theta, c, Q,\Phi^k)$ using block multiplication, yielding:
\begin{equation}\label{factor_model_1}
\begin{aligned}
\min_{\Theta,c, Q^TQ = I_{d+s},} \ell(\Theta, c, Q,\Phi^k) 
=\|X- \underbrace{[c, Q]
\left[
\begin{array}{ccc}
1 & 0 & 0 \\
0& I_d&\bf 0\\
0& \bf 0& \Theta^T
\end{array}
\right]}_{(a)}
\left[
{\bf 1}_n, {\Phi^k}^T, \Psi^T(\Phi^k) 
\right]^T
\|_{\rm F}^2.
\end{aligned}
\end{equation}
Due to the block product relationship between $Q$ and $\Theta^T$ in (a) and the orthonormal constraint imposed on $Q$, a closed-form solution does not exist concerning $\{\Theta, c, Q\}$. Nevertheless, an alternating perspective allows us to address this problem. In the following, we illustrate that the optimization problem with respect to any one of $\{\Theta, c, Q\}$, when fixing the other two remaining parameters, admits a closed-form solution.
For ease of notation, we denote $R(c):=X-c {\bf 1}_n^T$ and $M(\Theta):= [\Phi^T, \Psi^T(\Phi)\Theta]^T\in {\mathbb R}^{(d+s)\times n}$.



\zcolor{Note that the loss function  $\ell_\lambda(\Theta, c, Q, \Phi^k)$  includes an additional term $\lambda \|\Theta^T \Psi(\Phi^k)\|_{\rm F}^2$, compared to  $\ell(\Theta, c, Q, \Phi^k)$ . As a result, the main distinction in the alternating iterations lies in the updates for $\Theta$; the optimal solutions related to  $c$  and  $Q$  remain the same for both  $\ell_\lambda(\cdot)$  and  $\ell(\cdot)$  when other parameters are fixed.}


\subsubsection{Optimize $Q$ via Fixing $\{{\Theta},c, \Phi\}$\label{step1}}
Here, we derive the closed-form solution of $\ell(\Theta,c,Q,\Phi^k)$ when $\Theta$ and $c$ are held constant as $\Theta'$ and $c'$. By rearranging the objective function in \eqref{factor_model}, we can simplify the regression problem to optimize an orthonormal matrix $Q$ that effectively aligns the columns of $R(c')$ and $M(\Theta')$. This alignment is achieved by minimizing the subsequent objective function:
\begin{equation}\label{Othmin}
\min_{Q^TQ = I_{d+s}} \|R(c') - QM(\Theta')\|_{\rm F}^2.
\end{equation}
Since $Q$ is an orthonormal matrix with full column-rank, we have $\|QM(\Theta')\|_{\rm F}^2 = \|M(\Theta')]\|_{\rm F}^2 $. Consequently, we can eliminate constants and transform the problem into an equivalent form:
\begin{equation}\label{EquOthmin}
\hat{Q} = \arg\max_{Q^TQ=I_{d+s}} \langle R(c')M(\Theta')^T, Q \rangle.
\end{equation}
Since any orthonormal matrix $Q\in {\mathbb R}^{D\times (d+s)},$ can be expressed as the product of two orthonormal matrices as $YZ^T, Y\in {\mathbb R}^{D\times (d+s)}, Z\in {\mathbb R}^{(d+s)\times (d+s)}$ with $Y^TY = I_{d+s}$ and $Z^TZ = I_{d+s}$, we can rewrite \eqref{EquOthmin} as:
\begin{equation}\label{eqv}
(\hat{Y},\hat{Z})=\arg\max_{Y^T Y = I_{d+s}, Z^T Z = I_{d+s}} \langle R(c')M(\Theta')^T, Y Z^T \rangle. 
\end{equation}

It is evident that the optimal solution for \eqref{eqv} can be obtained through the application of Singular Value Decomposition (SVD)~\citep{SVD} on the matrix $R(c')M(\Theta')^T$. Specifically, the optimal objective value is $\sum_{i=1}^{d+s}\sigma_i(R(c')M(\Theta')^T)$ and the optimal solution is $\hat{Q}=\hat{Y}\hat{Z}^T$, where $\hat{Y}{\hat D}\hat{Z}^T$ is the singular value decomposition of $R(c')M(\Theta')^T$.
\subsubsection{Optimize $c$ via Fixing $\{Q, \Theta, \Phi\}$\label{step2}}

\zcolor{The problem in \eqref{factor_model} with respect to  $c$  is quadratic and thus convex, ensuring a global solution. By rearranging \eqref{factor_model} to emphasize  $c$  and fixing  $Q$ and $\Theta$  as  $Q'$  and $\Theta'$, respectively, we can express \eqref{factor_model} in the following equivalent form:}
\begin{equation}\label{min_x}
\min_{c}  \|X - Q'M(\Theta') - c {\bf 1}_n^T \|_{\rm F}^2.
\end{equation}
The above problem can be solved through its associated normal equation. Specifically, by computing the derivative with respect to $c$, the solution to \eqref{min_x} can be obtained by solving the normal equation
${\big (}X - Q'M(\Theta')-c {\bf 1}_n^T{\big)} {\bf 1}_n  = \bf 0$ which leads to a closed-form solution $\hat{c} = \frac{1}{n} (X- Q'M(\Theta')) {\bf 1}_n$.
\subsubsection{Optimize $\Theta$ via Fixing $\{c, Q, \Phi\}$\label{step3}}
For SQMF, the problem with respect to $\Theta$ is also quadratic and we can derive the closed-form solution for $\Theta$ through some projection manipulations. Since the spaces spanned by the columns of $U$ and $V$ are orthogonal, we can split the squared Frobenius norm of $\|R(c)-U\Phi-V\Theta^T\Psi(\Phi)\|_{\rm F}^2$ into:
\begin{equation}\label{xx}
    \min_{\Theta} \|V^T R(c)  - \Theta^T \Psi(\Phi)\|_{\rm F}^2 + \|U^T R(c)  -  \Phi\|_{\rm F}^2.
\end{equation}
Noticing that $\|U^T R(c)  -  \Phi\|_{\rm F}^2$ does not depend on $\Theta$, we can neglect this term and formulate \eqref{xx} equivalently as
\begin{equation}\label{opt_theta}
\min_\Theta \|V^T R(c)  - \Theta^T \Psi(\Phi)\|_{\rm F}^2.
\end{equation}
Since \eqref{opt_theta} is a regression problem, the optimizer of \eqref{opt_theta} can be found by solving the normal equation $\Psi(\Phi)\Psi(\Phi)^T \Theta - \Psi(\Phi)R(c)^T V = 0$ with respect to $\Theta$. Therefore, if $\Psi(\Phi)$ has full row-rank, we have the explicit solution for $\Theta$ as
\begin{equation}\label{theta}
 \hat{\Theta} =(\Psi(\Phi)\Psi(\Phi)^T)^{-1} \Psi(\Phi)R(c)^T V.
\end{equation}

For RSQMF, the procedure involving $\Theta$ can be executed similarly to derive the closed-form solution for
$
\min_\Theta \|V^T R(c)  - \Theta^T \Psi(\Phi)\|_{\rm F}^2+\lambda\|\Theta^T \Psi(\Phi)\|_{\rm F}^2
$ as follows:
\[
 \hat{\Theta} = \frac{1}{1+\lambda}(\Psi(\Phi)\Psi(\Phi)^T)^{-1} \Psi(\Phi)R(c)^T V.
\] 


\begin{table}[t]
\centering
\caption{The alternating algorithm for the regression problem. \label{alg:regression}}
\vspace{2mm}
\resizebox{\linewidth}{!}{
\begin{tabular}{l} \hline\hline
{\bf Input}: $X=[x_1,...,x_n]$ and the coordinate matrix $\Phi=[\tau_1,...,\tau_n]$; Dimensional parameters $d$ and $s$, \\which corresponds to the tangent space and subspace of the normal space, respectively.\\
{\bf Output}: The center: $c$; The orthonormal matrix $U$ and $V$ which span the tangent space \\
and the subspace of the normal space, respectively; The quadratic curvature parameter $\Theta\in {\mathbb R}^{\frac{d^2+d}{2}\times s}$. \\ \hline
(1).\hspace{2mm} Set the initial value of $\Theta$ as $\Theta_0=\bf 0$ and $Q$ as $Q_0=[I_{d+s},{\mathbf 0}]^T$.\\
(2).\hspace{8mm}  While $\|c_{t+1}-c_t\|_2^2+\|\Theta_{t+1}-\Theta_t\|_{\rm F}^2+\|Q_{t+1}Q^T_{t+1}-Q_t Q_t^T\|_{\rm F}^2\leq \epsilon$: \\
(3).\hspace{14mm} Update: $c_{t+1} = \frac{1}{n}(X-Q_t M(\Theta_t)) \bf 1$,  with $M(\Theta_t) = [\Phi^T, \Psi^T(\Phi)\Theta_t]^T$.\\
(4).\hspace{14mm}  Update: $Q_{t+1} := \arg\max_{Q^TQ=I_{d+s}} \langle R(c_{t+1})M(\Theta_t)^T, Q \rangle$.\\
(5).\hspace{14mm}  Update: $\Theta_{t+1} := (\Psi(\Phi)\Psi(\Phi)^T)^{-1} \Psi(\Phi)(R(c_{t+1}))^T V_{t+1}$, with $V_{t+1} = Q_{t+1}(:,[d+1:d+s])$.  \\
(6).\hspace{2mm} Return $(\Theta_{t+1},c_{t+1},U_{t+1}:=Q_{t+1}(:,[1:d]), V_{t+1}:=Q_{t+1}(:, [d+1,d+s]))$.\\
\hline
\end{tabular}}
\end{table}

In summary, by updating each of $Q,c$ and $\Theta$ as in \eqref{EquOthmin}, \eqref{min_x} and \eqref{theta} while fixing the remaining two parameters, we obtain a sequence of $\{Q_t,c_t,\Theta_t\}_{t=1}^{\infty}$. If the sequence converges, we use the limit point $(Q^*,c^*,\Theta^*)$ as the solution for the regression problem. The stopping condition is defined as the total variation of $c_t$, $\Theta_t$, and the spatial distance for the orthonormal matrix $Q$, with the maximum tolerance being $\epsilon$.
We present our iterative updating algorithm for the regression model in Table \ref{alg:regression}.

\subsection{Optimize $\ell(\Theta^k, c^k, Q^k,\Phi)$\label{sub2}}

In this subsection, we study how to optimize $\ell(\Theta^k, c^k, Q^k,\Phi)$ with respect to $\Phi$. Recall that the \zcolor{matrix-valued} function $\Psi(\Phi)$ is defined column-wise, and because the squared Frobenius norm is separable, solving $\min_\Phi \ell(\Theta^k, c^k, Q^k, \Phi)$ reduces to solving a projection problem for each column of $\Phi$ independently. Thus, we only need to optimize over each column variable, denoted here as $\tau$, to achieve the desired result, by the following problem:
\begin{equation}\label{proj}
\hat{\tau}_i = \arg\min_\tau \|x_i  - c- U\tau - V {\cal A}(\tau,\tau)\|_2^2.
\end{equation}
After solving \eqref{proj} for $i=1:n$, we can obtain $\hat{\Phi}$ by stacking each $\hat{\tau}_i$ into a matrix, i.e., $\hat{\Phi}=[\hat{\tau}_1, \hat{\tau}_2,...,\hat{\tau}_n]$.
Next, we further simplify \eqref{proj} into the summation of two loss functions on the space of $U$ and $V$. \zcolor{Since both of $U$ and $V$ are orthonormal matrices,} we have:
\begin{equation}\label{simiplified}
\begin{aligned}
   &\|x_i -  (V {\cal A} (\tau,\tau) + U \tau +c)\|_2^2\\
 =&\|P_U(x_i-c)-U\tau\|_2^2+\|P_V(x_i-c)-V {\cal A}(\tau,\tau)\|_2^2+\|(I-P_U-P_V)(x_i-c)\|_2^2 \\
  =&\|U (U^T (x_i-c)-\tau)\|_2^2+\|V (V^T(x_i-c)- {\cal A}(\tau,\tau))\|_2^2 +\|(I-P_U-P_V)(x_i-c)\|_2^2\\
 =&\|U^T (x_i-c)-\tau\|_2^2+\|V^T(x_i-c)- {\cal A} (\tau,\tau)\|_2^2+\|(I-P_U-P_V)(x_i-c)\|_2^2,
\end{aligned}
\end{equation}
where $P_U:=UU^T$ and $P_V:=VV^T$ are the projection matrices corresponding to two orthogonal spaces. By defining $\phi_i = U^T (x_i-c)\in {\mathbb R}^{d}, \psi_i = V^T(x_i-c)\in {\mathbb R}^{s}$ and eliminating the term that is not relevant with $\tau$, we can further simplify \eqref{simiplified} into
\begin{equation}\label{f_tau_op}
f_i(\tau) = \|\phi_i-\tau\|_2^2+\|\psi_i- {{\cal A}} (\tau,\tau)\|_2^2.
\end{equation}

\zcolor{In the following, we simplify notation by omitting the subscript  $i$  and denoting  $f_i(\tau)$  as  $f(\tau)$  where context permits. For the regularized version, noting that  $\|\Theta^T\Psi(\Phi)\|_{\rm F}^2$  is also a separable function, the problem with respect to each  $\tau$  reduces to:}
\begin{equation}\label{pen_version}
\min_{\tau}f_\lambda(\tau) =  f(\tau)+\lambda \|{\cal A}(\tau,\tau)\|_2^2.
\end{equation}


\zcolor{The optimization problems in \eqref{f_tau_op} and \eqref{pen_version} are nonlinear, presenting as quartic problems with respect to $\tau$. Due to the high-order term involving $\tau$, a closed-form solution is not feasible. Recognizing \eqref{f_tau_op} as a specific case of \eqref{pen_version} with $\lambda=0$, we focus on solving \eqref{pen_version}. We propose three approaches—Gradient Descent, Newton’s Method, and a Surrogate Method—to minimize \eqref{pen_version}. The theoretical properties of these methods are further analyzed in Section \ref{properties}.}

\subsubsection{Gradient and Newton's Methods}

Gradient descent~\citep{gradient_adaptive} is a commonly employed technique to minimize \zcolor{\eqref{pen_version}}. \zcolor{Starting} from an initial point $\tau_0$, the gradient descent proceeds as follows:
\[
\tau_{k+1} = \tau_k-\mu_k\nabla f_{\lambda}(\tau_k),
\]
where $\mu_k$ represents the step size. \zcolor{Practically}, we set $\mu_k = \frac{\mu_{k-1}}{2^k}$, with $k$ chosen as the smallest integer such that $f_\lambda(\tau_{k+1}) < f_\lambda(\tau_k)$. While gradient descent can reach linear convergence rates, further acceleration can be achieved by incorporating Hessian information, as done in Newton’s method.
\zcolor{Before presenting Newton's method, we first derive the gradient and Hessian of $f_\lambda(\tau)$ as follows:}
\begin{equation}\label{gH}
\begin{gathered}
    \nabla f_\lambda(\tau) = 2(\tau-\phi_i)+4 { A}_\tau^T((1+\lambda){\cal A}(\tau,\tau)-\psi_i) ,\\
 H_{f_\lambda}(\tau) = 2I_d+8(1+\lambda)A_\tau^T A_\tau+4{\cal A}^*({(1+\lambda)\cal A}(\tau,\tau)-\psi_i).
\end{gathered}
\end{equation}

\zcolor{Newton’s method~\citep{newton} can be understood as an approach that adjusts direction and step size using the inverse of the Hessian matrix. By utilizing the information in the Hessian, Newton’s method iteratively seeks the stationary point of  $f_i(\tau)$  through the following updates:}

\begin{equation}\label{Newton}
    \tau_{t+1} = \tau_t-H_{f_\lambda}^{-1}(\tau_t) \nabla f_\lambda(\tau_t).
\end{equation}

Unlike the gradient method, Newton’s method adaptively adjusts the step size through the Hessian matrix, removing the need for manual tuning. Instead of solving \eqref{pen_version} directly with standard optimization methods, a surrogate approach can be developed by reducing the problem’s order, allowing for an iterative solution through the surrogate method, as discussed in the following sections.

\subsubsection{Surrogate Method}
\zcolor{Here, we propose an alternative approach that leverages surrogate functions to reduce the problem’s order by introducing an auxiliary function. Specifically, we lower the order of  $f_\lambda(\tau)$  by incorporating an auxiliary function  $g_\lambda(\alpha,\beta)$, defined as follows:}
\begin{equation}\label{g12}
g_\lambda(\alpha,\beta) =\frac{1}{2} \|\phi_i-\alpha\|_2^2+\frac{1}{2} \|\phi_i-\beta\|_2^2+\|\psi_i- {\cal A} (\alpha,\beta)\|_2^2+\lambda \|{\cal A}(\alpha,\beta)\|_2^2.
\end{equation}
\zcolor{Due to the symmetry between $\alpha$ and $\beta$}, the function in \eqref{g12} satisfies $g_\lambda(\alpha,\beta) = g_\lambda(\beta,\alpha)$. Thus, if $(\alpha^*,\beta^*)$ is a minimizer of $g_\lambda(\alpha,\beta)$ with $\alpha^*=\beta^*$, then $\alpha^*$ also minimizes $f_\lambda(\tau)$. In the sequel, we elaborate on how to find the minimizer of $g_\lambda(\alpha,\beta)$ via the alternating minimization technique.

Starting from $\alpha_0=\beta_0=\phi_i$, we update $\{\alpha_n,\beta_n\}$ iteratively in the following manner:
\begin{equation}\label{min_tau_n}
\left\{
\begin{aligned}
\alpha_{n} =& \arg\min_{\alpha}g_\lambda(\alpha,\beta_{n-1}),\\
\beta_{n} =&\arg\min_{\beta}g_\lambda(\alpha_{n},\beta).
\end{aligned}
\right.
\end{equation}

\zcolor{Through the iterations in \eqref{min_tau_n}, we generate a sequence of $\{(\alpha_n,\beta_n), i=1,2,\cdots\}$.
Once this sequence converges, with $\lim_{n\rightarrow \infty}\alpha_n=\lim_{n\rightarrow\infty}\beta_n$, the limit point serves as the solution to the problem $\min_\tau f_\lambda(\tau)$.}

This surrogate method can be implemented efficiently as iterates in \eqref{min_tau_n} admit closed-form solutions. Recall that $g_\lambda(\alpha,\beta_{n-1})$ yields a form as
\begin{equation}\label{g_lambda}
g_\lambda(\alpha,\beta_{n-1}) =\frac{1}{2} \|\phi_i-\alpha\|_2^2+\frac{1}{2} \|\phi_i-\beta_{n-1}\|_2^2+\|\psi_i- {\cal A} (\alpha,\beta_{n-1})\|_2^2+\lambda \|{\cal A}(\alpha,\beta_{n-1})\|_2^2.
\end{equation}
\zcolor{
Thus, minimizing \eqref{g_lambda} results in a quadratic function in terms of $\alpha$, allowing us to obtain the following closed-form solution:}
\begin{equation}\label{alphan}
\alpha_n = (2(1+\lambda){A}^T_{\beta_{n-1}}{A}_{\beta_{n-1}}+I_d)^{-1}(2{\cal A}^*(\psi_i)\beta_{n-1}+\phi_i).
\end{equation}
\zcolor{Recognizing that $g(\alpha,\beta)$ is symmetric, minimizing $g(\alpha_{n},\beta)$ with respect to $\beta$ can be addressed similarly as follows:}
\begin{equation}\label{betan}
\beta_n = (2(1+\lambda){A}^T_{\alpha_{n}}{A}_{\alpha_{n}}+I_d)^{-1}(2{\cal A}^*(\psi_i)\alpha_n+\phi_i).
\end{equation}

Till now, we have finished the algorithmic section for solving the subspace-contained quadratic matrix factorization model. In what follows, we will demonstrate that $\{(\alpha_n,\beta_n)\}$ converges  by showing that $g_i(\alpha,\beta)$ is a strongly convex function \zcolor{over ${\cal S}_\gamma \times {\cal S}_\gamma$} and the sequence $\{(\alpha_n,\beta_n)\}$ is contained in \zcolor{${\cal S}_{\gamma}\times {\cal S}_{\gamma}$} under certain conditions.

\subsection{Comparisons of Optimization Strategies}
\begin{figure}[t]
{\hspace{-0.1\linewidth}
\includegraphics[width=1.2\linewidth]{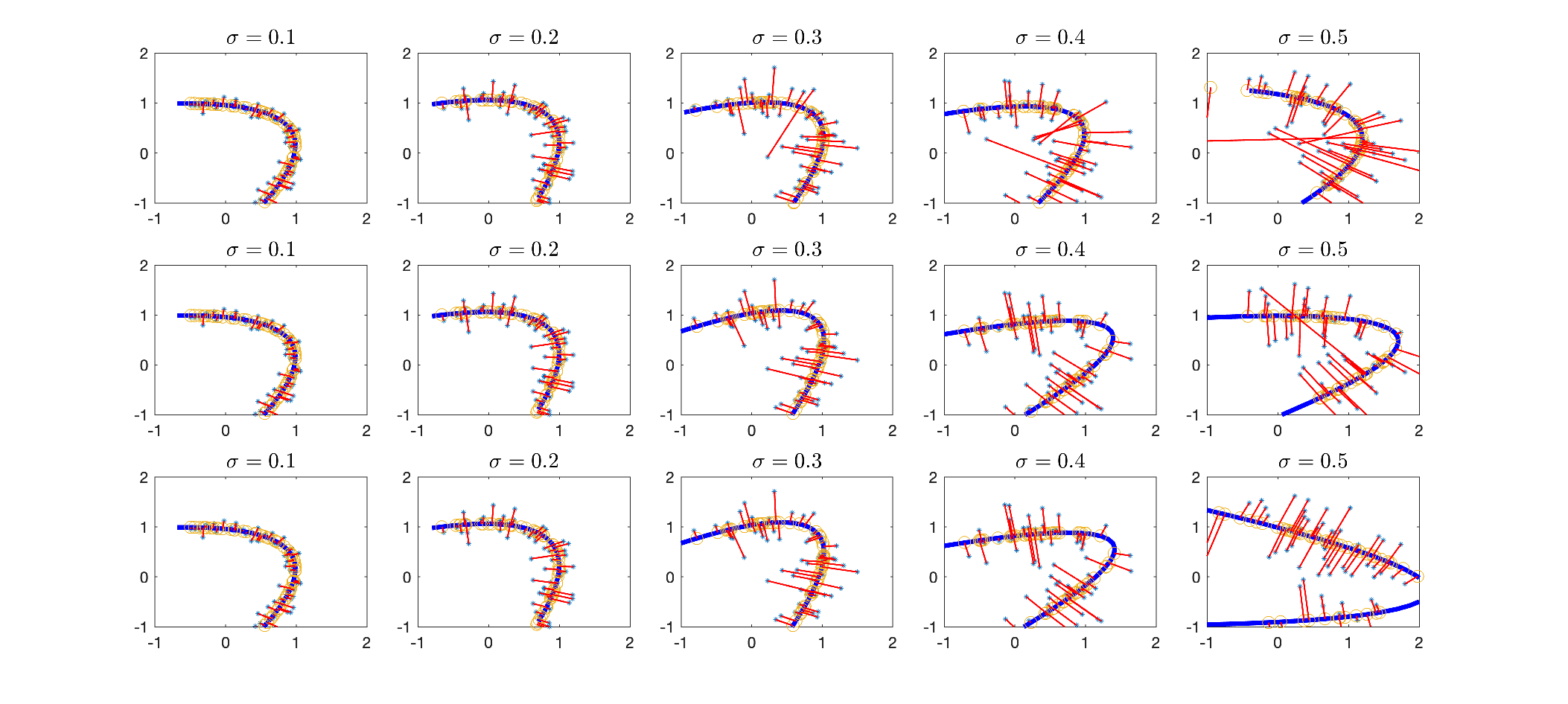}}
\vspace{-15mm}
\caption{The illustration depicts a fitted quadratic curve and the process of projecting noisy data onto the curve while minimizing the $\ell_2$ distance across various noise levels represented by $\sigma$. We demonstrate the performance of the Newton's method in the first row, the gradient algorithm in the second row, and the surrogate method in the third row.}
\label{fig:fit_result}
\end{figure}

In this subsection, we examine the performance of three different approaches: Gradient, Newton, and the Surrogate method for solving the projection subproblem of \eqref{f_tau_op}. For the sake of a fair comparison, we employ the same alternating strategy to solve the regression problem in the outer iteration and compare the convergence behavior corresponding to the different strategies for the inner problem. The data $\{(x_i, y_i), i=1, \ldots, n\}$ is generated as $x_i = \cos(t_i) + \epsilon_{x}(t_i)$, $y_i = \sin(t_i) + \epsilon_{y}(t_i)$, where $\{t_i,i=1,\ldots,n\}$ are sampled equally in $[-\frac{\pi}{3}, \frac{2\pi}{3}]$, and $\epsilon_{x}(t), \epsilon_{y}(t) \sim \mathcal{N}(0, \sigma^2)$.

\begin{figure}[t]
\hspace{-0.06\linewidth}
\includegraphics[width=1.12\linewidth]{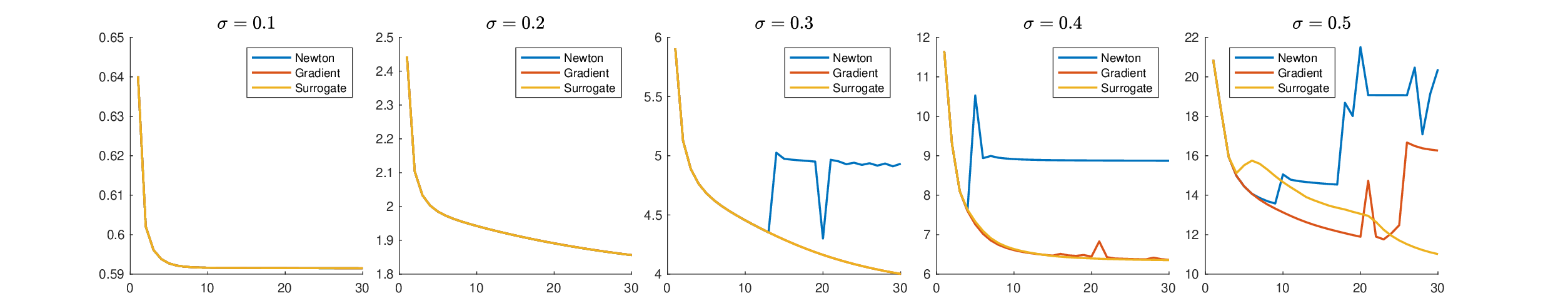}
\vspace{-10mm}
\caption{The comparison of the variation of the function value $\|X - Q^k M(\Theta^k) - c^k {\bf 1}^T \|_{\rm F}^2$ changes with the iteration steps $k$ for three different approaches: Gradient, Newton and Surrogate method. 
\label{fig:iter_steps}}
\vspace{-3mm}
\end{figure}

We illustrate the fitting result with the blue curve and also demonstrate how the noisy point is projected onto the fitted curve with a red line. Additionally, we plot how the fitting error, measured in the squared Frobenius norm $\|X - Q M(\Theta) - c \mathbf{1}^T \|_{\rm F}^2$, changes with the iteration steps in Figure~\ref{fig:iter_steps}. The total CPU time cost is reported in Table \ref{tab:cputime}. From Figure~\ref{fig:iter_steps} and Table \ref{tab:cputime}, we can draw the following conclusions:
\begin{itemize}[leftmargin=*]
\item When the noise level is small, all three approaches exhibit a monotonically decreasing trend in the function value of the objective. However, as the noise level increases, this monotonically decreasing phenomenon no longer exists, as seen in $\sigma=0.5$ in Figure~\ref{fig:iter_steps}.
\item 
Compared with Newton's convergence curve, the gradient and surrogate methods demonstrate stronger robustness with the increasing value of $\sigma$. We observe that Newton's method exhibits unstable convergence behavior for $\sigma=0.3$ and $0.4$, while the gradient and surrogate methods maintain stability.
\end{itemize}

The unsatisfactory convergence behavior of the three optimization methods is caused by the uncertainty regarding the convexity of $f_i(\tau)$ for each $i$ as $\sigma$ increases. Ensuring convergence behavior requires us to provide a theoretical justification for employing these methods to solve the projection problem \eqref{f_tau_op} in the subsequent sections.

\begin{table}[t]
\centering
\caption{The comparison of the CPU time for solving the optimization  problem \eqref{factor_model} of three approaches under different noise level.\label{tab:cputime}}
\vspace{2mm}
\resizebox{0.6\linewidth}{!}{
\begin{tabular}{c|ccccc} \toprule
  { Methods}   & $\sigma = 0.1$ & $\sigma=0.2$ & $\sigma=0.3$ & $\sigma=0.4$ & $\sigma = 0.5$\\ \hline
  Newton  &  0.04  &  0.04  &   0.07 &   0.04  &  0.04 \\
  Gradient  &  0.13  &  0.13  &  0.13  &  0.16  &  0.16 \\
  Surrogate  &  0.05  &  0.04  &  0.06  &  0.21  &  0.22 \\ \bottomrule
\end{tabular}}
\end{table}

\subsection{Complexity Analysis}
In this section, we present a detailed analysis of the computational complexity of the RSQMF method. Given that RSQMF consists of two main components—regression and projection—we provide separate analyses for each step.
 \begin{table}[h]
 \centering
\caption{Computational complexity for updating $\Theta$ till the current term.\label{Complexity_Theta}}
\vspace{2mm}
 \resizebox{\linewidth}{!}{
 \begin{tabular}{c|c|c|c|c} \hline\hline
 Terms &   $\Psi(\Phi)\Psi(\Phi)^T$  & $(\Psi(\Phi)\Psi(\Phi)^T)^{-1}$  & $(\Psi(\Phi)\Psi(\Phi)^T)^{-1}\Psi(\Phi)$ &$(\Psi(\Phi)\Psi(\Phi)^T)^{-1}\Psi(\Phi) R(c)^T V$ \\\hline
 Complexity & $O(\
nd^4)$   & $O(nd^4+d^6)$ & $O(nd^4+d^6+d^4n)$& $O(nd^4+d^6+d^2n D +d^2D s)$ \\ \hline
\end{tabular}}
 \caption{Computational complexity for updating $c$ and $Q$ till the current term.\label{Complexity_cQ}}
 \vspace{2mm}
 \begin{tabular}{c|c|c|c}\hline\hline
   Terms  &$M(\Theta_t)$ & $(X-Q_t M(\Theta_t)) \bf 1$ & $R(c)M(\Theta)^T$ \\ \hline
  Complexity &  $O(n s d^2)$ & $O(n s d^2+nDs+nDd)$ & $O(nsd^2+Dnd+Dns)$\\ \hline
 \end{tabular}
\caption{Computational complexity for updating $\tau$ for Gradient, Newton and Surrogate Method.\label{Complexity_tau}}
\vspace{2mm}
 \begin{tabular}{c|c|c|c} \hline\hline
Terms &  $\nabla f_\lambda(\tau)$    & $H_{f_\lambda}^{-1}(\tau)$ & Surrogate Iteration in \eqref{alphan} or \eqref{betan}\\ \hline
Complexity &   $O( s d^2)$   &$O(sd^2+d^3)$ & $O(sd^2+d^3)$ \\ \hline
 \end{tabular}
 \end{table}

The regression step involves updating the matrices  $Q$,  $c$, and  $\Theta$  using equations \eqref{eqv}, \eqref{min_x}, and \eqref{theta}, respectively. This process includes the computation of  $\Psi(\Phi)$, the Moore-Penrose inverse  $(\Psi(\Phi)\Psi(\Phi)^T)^{\dagger}$, and several matrix multiplications. Given that  $X \in \mathbb{R}^{D \times n}$  and  $\Psi(\Phi)$ has $O(d^2)$  rows and $n$  columns, the computational complexity for updating  $\Theta$ is  $O(d^4n + d^6 + Dnd^2 + Dd^2s)$.


The construction of  $M(\Theta) := [\Phi^T, \Psi^T(\Phi)\Theta]^T \in \mathbb{R}^{(d+s) \times n}$  involves a matrix computation of  $\Psi^T(\Phi)\Theta$ , with a complexity of  $O(nsd^2)$ . Since  $Q_t \in \mathbb{R}^{D \times (d+s)}$  and  $M(\Theta_t) \in \mathbb{R}^{(d+s) \times n}$ , the complexity for computing  $Q_t M(\Theta_t)$  is  $O(Ddn + Dsn)$. Therefore, the overall computational complexity for updating  $c$  is  $O(nsd^2 + nDs + nDd)$ .


For the computation of  $R(c)M(\Theta)^T$  where  $R(c) \in \mathbb{R}^{D \times n}$ , the complexity is  $O(nsd^2 + Dnd + Dns)$. Additionally, updating  $Q$  requires performing a partial SVD corresponding to the previous  $d+s$  eigenvectors, which costs  $O(dD^2 + sD^2)$  for a  $D \times D$  matrix. Hence, the total computational complexity for updating  $Q$  is  $O(nsd^2 + Dnd + Dns + dD^2 + sD^2)$. 


In the projection step, we solve equation \eqref{proj} iteratively for each of the  $n$  tasks. Each task is equivalent to solving problem \eqref{pen_version} using either gradient, Newton, or surrogate methods. This requires computing the gradient  $\nabla f_{\lambda}(\tau)$, the inverse Hessian  $H_{f_\lambda}^{-1}(\tau)$, and performing matrix multiplications, followed by executing steps like \eqref{Newton} or \eqref{alphan}. In each projection step, the complexity for the gradient method is  $O(sd^2)$ , while for the Newton and Surrogate methods, it is  $O(d^3 + sd^2)$. 

In summary, Table \ref{Complexity_Theta}, Table \ref{Complexity_cQ}, and Table \ref{Complexity_tau} provide a detailed overview of the computational complexities for updating  $\Theta$, $c$, and $Q$ in the regression step, as well as for the gradient, Newton, and surrogate methods in the projection step.


\subsection{Theoretical Properties} \label{properties}

In this section, we will elucidate the rationale for solving $\min_\tau f_i(\tau)$ using Newton's iteration and the surrogate method in \eqref{min_tau_n}, leveraging the strong convexity of $g_i(\alpha,\beta)$ within a local region of ${\cal S}_\gamma:=\{\alpha\in {\mathbb R}^d,\|\alpha\|_2\leq \gamma \}$ under certain conditions. The strong convexity can be verified by examining the Hessian of $g_i(\alpha,\beta)$. For simplicity, we study the case for the non-regularized objective function by setting $\lambda=0$ in $f_\lambda(\cdot)$.

\begin{lemma}\label{convex_theorem}
Let $\gamma>0$ and $A_k$ be the $k$-th slice of $\cal A$ for $k = 1,...,D-d$. If $\mathfrak b = \max_k \sigma_1(A_k)$ satisfies
\begin{equation}\label{bound}
(D-d)\gamma^2{\mathfrak b}^2 - \frac{1}{2}\|\psi_i\|_1\mathfrak b \leq 1/8,
\end{equation}
then, $\lambda_{min}(H_{g_i}(\alpha,\beta))>\frac{1}{2}$ for all $\tau,\eta \in {\cal S}_\gamma$. In other words, $g_i(\alpha,\beta)$ is strongly convex over the region ${\cal S}_\gamma \times {\cal S}_\gamma$.
\end{lemma}

The proof of Lemma 1 is postponed to the Appendix. We can draw two conclusions from observing \eqref{bound}:
First, $g_i(\alpha,\beta)$ is strongly convex over ${\cal S}_\gamma \times {\cal S}_\gamma$ as long as the maximum curvature, measured by $\mathfrak b$, is not very large. Specifically, $\mathfrak b$ should be within the range $\left(0, \frac{\|\psi_i\|_1+\sqrt{\|\psi_i\|_1^2+2(D-d)\gamma^2}}{4(D-d)\gamma^2}\right)$. Second, for any $\mathfrak b$, regardless of its magnitude, $g_i(\alpha,\beta)$ is strongly convex over $\cal S_\gamma\times \cal S_\gamma$ as long as $\gamma\leq \sqrt{\frac{1+4\|\psi_i\|_1\mathfrak b}{8(D-d)\mathfrak b^2}}$.

The strong convexity of ${\cal S}_\gamma\times {\cal S}_\gamma$, along with the symmetric property of $g_i(\alpha,\beta)$, implies that there exists only one optimal solution as long as $\gamma$ satisfies the condition in Lemma \ref{convex_theorem}. This leads to our following theorem:

\begin{theorem}\label{converge}
\zcolor{
Suppose the conditions in Lemma 1 are satisfied and  that $\{(\alpha_s,\beta_s)\}$ generated by \eqref{min_tau_n} resides within ${\cal S}_\gamma\times {\cal S}_\gamma$, then, $\{(\alpha_s, \beta_s)\}$ converges as $s\rightarrow +\infty$ and $\lim_s \alpha_s=\lim_s \beta_s$. Furthermore, we have $\lim_s g(\alpha_s,\beta_s) = f(\tau^*)$, where $\tau^*:=\lim_s \alpha_s=\lim_s \beta_s$.}
\end{theorem}
The proof of Theorem \ref{converge} is postponed to the Appendix. Theorem \ref{converge} requires that the sequence ${(\alpha_s,\beta_s)}$ falls into ${\cal S}_\gamma\times {\cal S}_\gamma$, which is difficult to check. In what follows, we provide an upper bound for the sequence $\{(\alpha_n,\beta_n)\}$ under mild conditions, such as $\|{\cal A}^*(\psi_i)\|_{\rm op} < 1$. 

Despite the absence of a closed-form solution for $g_i(\alpha, \beta)$, valuable insights into the norm of the sequence $\{(\alpha_n,\beta_n)\}$ can be gleaned from the iterations under certain mild conditions.

\begin{proposition}\label{norm}
For any $\epsilon > 0$ and $\|{\cal A}^*(\psi_i)\|_{\rm op} < 1$, there exists $K_0(\epsilon)$ 
such that the sequence $\{\alpha_n\}$,$\{\beta_n\}$ obtained from \eqref{min_tau_n} satisfies $\max\{\|\alpha_n\|_2,\|\beta_n\|_2\} \leq \frac{\|\phi_i\|_2}{1-2\|{\cal A}^*(\psi_i)\|_{\rm op}}$ for $n>K_0(\epsilon)$.
\end{proposition}

\zcolor{The proof of Proposition \ref{norm} is deferred to the Appendix. Based on Proposition \ref{norm}, we can deduce that the upper bound for the norm at the convergence point is solely determined by  $\|\phi_i\|_2$  and  $\|{\cal A}^*(\psi_i)\|_{\text{op}}$. In the limiting case where  ${\mathfrak b} \to 0$, the problem reduces to a linear projection, with  $\alpha^*$ as the projection of  $x_i$  onto the tangent space, specifically  $\alpha^*$ = $\phi_i$.}


\begin{corollary}\label{coro}
    If the conditions in Proposition \ref{norm} \zcolor{hold}, and $\frac{\|\phi_i\|_2}{1-2\|{\cal A}^*(\psi_i)\|_{\rm op}} \leq \sqrt{\frac{1+4\|\psi_i\|_1\mathfrak b}{8(D-d)\mathfrak b^2}}$, the sequence $\{(\alpha_n,\beta_n)\}$ generated from \eqref{min_tau_n} converges, with $\lim_n \alpha_n = \lim_n \beta_n$.
\end{corollary}

The proof of Corollary \ref{coro} is straightforward, as we \zcolor{can validate} that \zcolor{${\cal S}_\gamma\times {\cal S}_{\gamma}$} is convex and contains the sequence $\{(\alpha_n, \beta_n)\}$ for sufficiently large $n$. Corollary \ref{coro} follows since the function $g_i(\cdot, \cdot)$, when restricted to ${\cal S}_\gamma \times {\cal S}_\gamma$, attains a unique optimal solution.

\section{Convergence Properties}
This section examines the convergence properties for both the inner and outer iterative updates. Specifically, we outline conditions under which the convergence point satisfies the Karush-Kuhn-Tucker (KKT) conditions~\citep{convex optimization} for the corresponding optimization problem. The results are presented here, with detailed proofs for Theorems \ref{converge-in} and \ref{converge-out} provided in the Appendix.

\zcolor{First, we analyze the convergence of the sequence when solving the regression problem (inner iteration)  $\min_{Q, c, \Theta} \ell(Q, c, \Theta)$  with  $\Phi$  fixed. We can conclude that any accumulation point  $(Q^*, c^*, \Theta^*)$  will satisfy the KKT conditions, assuming the sequences  $\{c_t, Q_t\}$  have accumulation points and that  $\Psi(\Phi)$  is full row-rank.}

\begin{theorem}\label{converge-in}
If there exists \zcolor{a} subsequence $\{c_{t_j},Q_{t_j}\}_{j=1}^{\infty}$ such that $\lim_j c_{t_j} = c^*,\lim_j Q_{t_j}=Q^*$, and if the smallest eigenvalue of $\lambda_{\min}(\Psi(\Phi)\Psi(\Phi)^T)>0$, then the sequence $\{\Theta_{t_j}\}_{j=1}^{\infty}$ also converges, with $\Theta^* = \lim_j \Theta_{t_j}$. Furthermore, if the singular values of $R(c^*)M(\Theta^*)^T$ are distinct, the accumulation point $\{Q^*,c^*,\Theta^*\}$ satisfies the first-order Karush-Kuhn-Tucker (KKT) conditions for the problem in \eqref{factor_model}. 
\end{theorem}


\zcolor{Second, we examine the convergence properties of the algorithm for the subspace-constrained quadratic matrix factorization problem (outer iteration). To facilitate this analysis, let us define the set  ${\cal B}_i := \{\tau \mid \sigma_{\max} (2A_\tau^T A_\tau + {\cal A}^*( {\cal A}(\tau, \tau) - \psi_i )) < 1/4 \}$ , where  $\psi_i$  is parameterized by  $x_i, c, Q$, and  $\cal A$ is the tensor form of  $\Theta$. We can then deduce that  $(\Theta^*, c^*, Q^*, \Phi^*)$  satisfies the Karush-Kuhn-Tucker (KKT) conditions of  $\ell(\Theta, c, Q, \Phi)$ , where  $(\Theta^*, c^*, Q^*)$  is any accumulation point of  $(\Theta_{j_k}, c_{j_k}, Q_{i_k})$, and  $\Phi^*$  is uniquely determined by $(\Theta^*, c^*, Q^*)$.}



\begin{theorem}\label{converge-out}
If there exists a sub-sequence $\{\Theta_{j_k},c_{j_k},Q_{j_k},k=1,2,...,\infty\}$ such that 
$
(\Theta^*,c^*,Q^*) = \lim_{k\rightarrow \infty} (\Theta_{j_k},c_{j_k},Q_{j_k})
$
and  ${\cal B}_1\times {\cal B}_2\times...\times {\cal B}_n$ is a convex set depending on $\{x_i,i=1,2,...,n\}$ and $c^*,Q^*,\Theta^*$.
Then, $\{\Theta_{j_k},c_{j_k},Q_{j_k},\Phi_{j_k},k=1,2,...,\infty\}$ also converges, and the accumulation point $(\Theta^*,c^*,Q^*,\Phi^*)$ satisfies the Karush-Kuhn-Tucker (KKT) conditions of $\ell(\Theta,c,Q,\Phi)$.
\end{theorem}


\section{Applications}\label{application}
\zcolor{
The subspace-constrained quadratic matrix factorization model fulfills two crucial roles: First, it assists in revealing the complex high-dimensional geometric structure inherent in a dataset. This encompasses tasks like refining noisy manifold-distributed data and inferring the tangent space at specific locations. Second, it facilitates the generation of a low-dimensional embedding representation within the latent space, which can be leveraged to elucidate the underlying patterns driving the variation trends in high-dimensional data.}
\subsection{Refining Noisy Manifold-Structured Data}
The model RSQMF is adept at denoising manifold-structured data by learning a high-dimensional quadratic surface in a local region of ${\cal N}_r(x_i)$. Through the formulation of a local quadratic fitting model $f(\tau) = c + U\tau + V\cal A(\tau,\tau)$ in the vicinity ${\cal N}_r(x_i)$, discrete data information is generalized into a continuous quadratic manifold. Afterwards, we can project $x_i$ onto the range of $f(\tau)$ via a nonlinear least squares problem as  $\hat{x}_i=\arg \min_{x=f(\tau)} \|x-x_i\|_2^2$. Therefore, we can measure the reconstruction error via the mean squared error over all the samples:
\[
{\cal F}_e = \frac{1}{n} \sum_{i=1}^n \|\hat{x}_i-x^*_i\|_2^2,
\]
where $\hat{x}_i$ and $x_i^*$ represent the projection of $x_i$ onto $f(\tau)$ and the underlying truth for $x_i$, respectively.
\subsection{Tangent Space Estimation}
\zcolor{The RSQMF model enables effective tangent space estimation. By fitting the data with RSQMF, we derive a representation  $f(\tau) = c + U\tau + V{\cal A}(\tau,\tau)$, where the columns of  $U$  define the tangent space at point  $c$. For an unknown point  $x$  near  $c$, we estimate the tangent space at  $x$  by locally adapting the tangent space defined by $U$. This adaptive approach allows the model to capture variations in the tangent space within the local neighborhood of  $c$, enhancing the precision of tangent space estimation.}

 \zcolor{The reasoning behind this adjustment is as follows}: by taking the derivative of $f(\tau)$ with respect to $\tau$, we find $\nabla f(\tau) = U+2V{\cal A}(\tau)$. Consequently, the tangent space at $\tau$ is shaped by altering the column space of $U$ with the column space of $2V{\cal A}(\tau)$. Given that $U+2V{\cal A}(\tau)$ deviates from being an orthonormal matrix, we employ QR decomposition to yield ${\cal Q R}: = U+2V{\cal A}(\tau)$, where $\cal Q$ is an orthonormal matrix and $\cal R$ is an upper-triangle matrix. We then utilize the columns of $\cal Q$ as the basis for the adjusted tangent space at $x$. In addressing a tangent space estimation problem for a set of points $\{x_i,i=1,...,n\}$, we adopt the average distance between the estimated tangent space and the true underlying space as a criterion for evaluating the performance of the tangent estimator. Specifically, we define
\[
{\cal T}_e = \frac{1}{n}\sum_{i=1}^n \|P_{{\cal Q}_i}-P_i^{*}\|_{\rm F}^2,
\]
where  $P_{{\cal Q}_i}={\cal Q}_i{\cal Q}_i^T$ and $P_i^{*}=U_i^*{U_i^*}^T$ represent the estimated and the underlying truth projection matrices for the tangent space at $x_i$, respectively.

\subsection{Enhanced Representation Capability}
The SQMF or the RSQMF model exhibits enhanced representation capabilities, expressing data through the function $f(\tau) = c + U\tau + V\mathcal{A}(\tau,\tau)$. The linear term $c + U\tau$ captures the principal signal, while the remaining part $V\mathcal{A}(\tau,\tau)$ acts as a fine adjustment. The inclusion of the nonlinear term enables RSQMF to capture nonlinear signals, making it more versatile than traditional methods. In Section \ref{handwritten}, we will demonstrate the superiority of RSQMF over related matrix factorization methods in reconstructing handwritten images with a latent representation in $\mathbb{R}^3$ using the MNIST dataset. Additionally, the embeddings derived from RSQMF exhibit a more interpretable representation in the latent space compared to traditional matrix factorization methods.

\section{Numerical Experiments} \label{experiment}
This section is structured into four key parts. First, we present a comparative analysis of RSQMF against five related algorithms, emphasizing its performance in manifold denoising (or fitting) and tangent space estimation. We also provide a detailed comparison of the computational costs associated with each method, measured in terms of CPU time. In the following part, we showcase the quadratic factorization model’s capability to capture variation trends in high-dimensional spaces by training RSQMF within a local neighborhood. Third, we apply RSQMF to a specially constructed two-class dataset, comprising randomly selected samples of the digits `4’ and `9’ from the MNIST dataset~\cite{mnist}, demonstrating its superior image reconstruction abilities compared to other matrix factorization methods such as \zcolor{ NMF~\citep{mf2}, Semi-NMF~\citep{smf}, MMF~\citep{ZhangZhao}, MF~\citep{MF}, and RQMF~\citep{Zhai}}. Lastly, we illustrate RSQMF’s effectiveness in denoising high-dimensional images, where it produces clean representations by utilizing information from neighboring data points. The code supporting this paper is available at https://github.com/zhaizheng/SQMF.

\subsection{Experiments with Synthetic Data}
\begin{figure}[t] 
   \centering  \includegraphics[width=\linewidth]{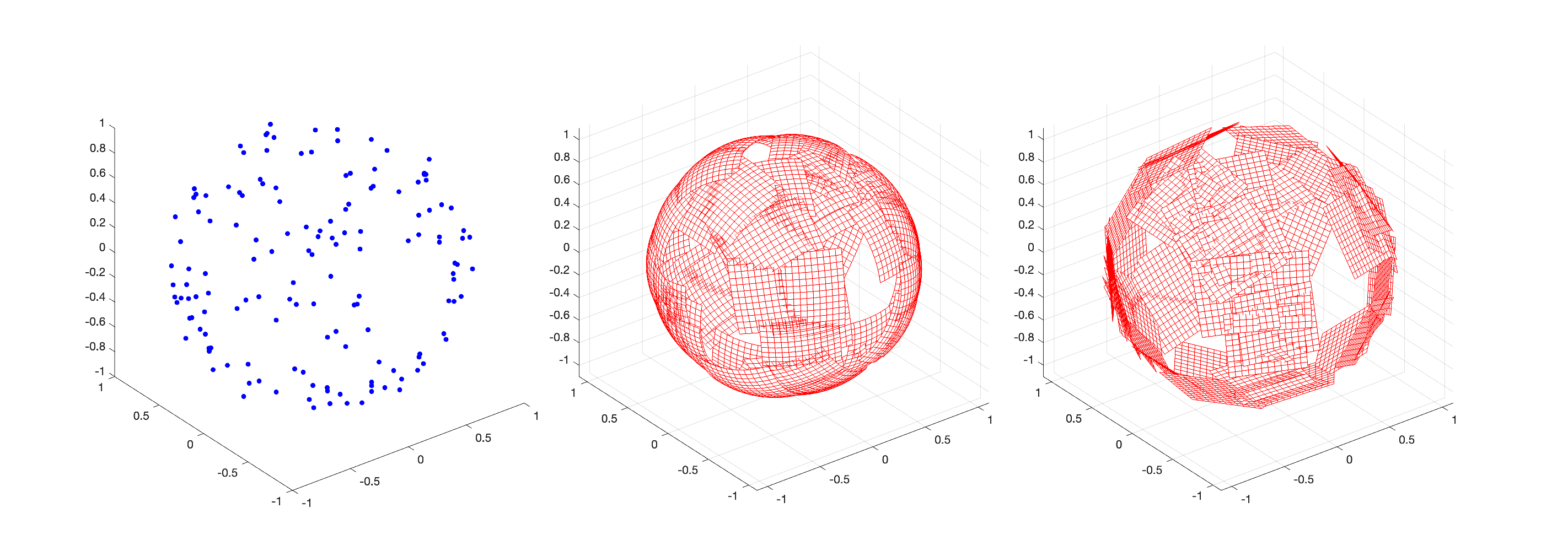} 
   \vspace{-10mm}
   \caption{ Visualization of the function $f(\tau)$ and its associated tangent space via fitting the samples in the vicinity of a two-dimensional sphere..
   Left: Synthetic noisy data approximately distributed on a 2-dimensional sphere. Middle: \zcolor{The illustration of the estimated $f(\tau)$ by fitting the data in each local region.}
   Right: Visualization of the tangent space.\label{demo}}
\end{figure}

In this section, we perform a comparative analysis of RSQMF, \zcolor{MLS~\citep{mls}, LOG-KDE~\citep{msf}, MFIT~\citep{mfit}, LPCA~\citep{lpca}, and RQMF~\citep{Zhai} }using a synthetic spherical experiment. These methods are evaluated based on their ability to simultaneously denoise the manifold and estimate the tangent space, providing an assessment of their performance in both tasks.
The data is generated through the following process: $x_i = \tilde{x}_i + \epsilon_i$, where $\{\tilde{x}_i\in {\mathbb R}^3,i=1,\cdots, n\}$ are uniformly distributed on a 2-dimensional sphere, and $\{\epsilon_i, i=1,...,n\}$ are i.i.d. Gaussian noise with $\epsilon_i \sim \mathcal{N}({\bf 0}, \sigma^2 {I}_3)$, as depicted in the leftmost diagram in Figure \ref{demo}. We employ the quadratic model proposed in our paper to learn the latent spherical manifold and its tangent space from the noisy dataset. It is worth mentioning that, in this specific case, the tangent space yields a rather simple estimator in the form of $P_{\tilde{U}_i} = I_D -  \tilde{x}_i {\tilde{x}_i}^T/\|\tilde{x}_i\|_2^2$. 
\zcolor{This specific form of the tangent space allows us to conveniently evaluate recovery performance from both manifold and tangent space perspectives simultaneously.}
The recovery performance is illustrated in the middle and right diagrams in Figure \ref{demo}.

\begin{table}[t]
\centering
\caption{Comparison of RSQMF with five related methods measured by ${\cal F}_e$ and ${\cal T}_e$ on synthetic spherical data, with varying neighborhood sizes $K$ and noise levels $\sigma$.\label{synthetic}}
\vspace{2mm}
\resizebox{0.98\linewidth}{!}{
\begin{tabular}{c|c|ccccccc|ccccccc}\toprule
& & \multicolumn{7}{c|}{${\cal F}_e$}& \multicolumn{7}{c}{${\cal T}_e$} \\\hline
Methods&\small {$\sigma$}{\large \textbackslash}{$K$} 
& 22 & 26 & 30 & 34 & 38 & 42 & 46 
& 22 & 26 & 30 & 34 & 38 & 42 & 46 
\\ \hline
\multirow{3}{*}{RSQMF} 
& 0.06 & 0.0007 & 0.0008 & 0.0008 & 0.0009 & 0.0010 & 0.0012 & 0.0014 & 0.0034 & 0.0037 & 0.0043 & 0.0052 & 0.0066 & 0.0079 & 0.0097 \\
&0.08 & 0.0013 & 0.0013 & 0.0014 & 0.0014 & 0.0016 & 0.0018 & 0.0020 & 0.0047 & 0.0047 & 0.0053 & 0.0059 & 0.0069 & 0.0074 & 0.0085 \\
&0.10 
& 0.0019 & 0.0019 & 0.0019 & 0.0020 & 0.0021 & 0.0022 & 0.0025 & 0.0072 & 0.0066 & 0.0072 & 0.0083 & 0.0100 & 0.0122 & 0.0138 \\ \hline
\multirow{3}{*}{MLS} 
&0.06 & 0.0008 & 0.0009 & 0.0010 & 0.0012 & 0.0015 & 0.0020 & 0.0026 & 0.0158 & 0.0168 & 0.0177 & 0.0173 & 0.0168 & 0.0166 & 0.0174 \\
&0.08 & 0.0013 & 0.0014 & 0.0015 & 0.0017 & 0.0020 & 0.0024 & 0.0029 & 0.0194 & 0.0220 & 0.0243 & 0.0244 & 0.0220 & 0.0212 & 0.0222 \\
&0.10 & 0.0020 & 0.0020 & 0.0020 & 0.0021 & 0.0023 & 0.0026 & 0.0031 & 0.0252 & 0.0283 & 0.0267 & 0.0263 & 0.0221 & 0.0263 & 0.0266 \\ \hline
\multirow{3}{*}{\zcolor{LOG-KDE}} 
&0.06& 0.0161 & 0.0273 & 0.0485 & 0.0619 & 0.0933 & 0.0974 & 0.1197 & 2.6562 & 2.4458 & 2.2212 & 2.0194 & 1.8457 & 1.6962 & 1.5822 \\
&0.08 & 0.0171 & 0.0310 & 0.0455 & 0.0720 & 0.0971 & 0.1017 & 0.1005 & 2.8342 & 2.5540 & 2.1937 & 1.7409 & 1.4936 & 1.3412 & 1.2789 \\
&0.10 & 0.0156 & 0.0296 & 0.0548 & 0.0908 & 0.1179 & 0.1548 & 0.1779 & 2.7181 & 2.4683 & 2.0651 & 1.7954 & 1.5867 & 1.4913 & 1.3757 \\ \hline
\multirow{3}{*}{MFIT} 
& 0.06 & 0.0275 & 0.0495 & 0.0871 & 0.1747 & 0.2240 & 0.2429 & 0.2869 & 2.8358 & 2.7366 & 2.6293 & 2.4995 & 2.4235 & 2.2897 & 2.2252 \\
& 0.08 & 0.0231 & 0.0337 & 0.0430 & 0.0509 & 0.0643 & 0.0899 & 0.0944 & 2.8854 & 2.8582 & 2.8236 & 2.7628 & 2.6508 & 2.4897 & 2.3754 \\
& 0.10 & 0.0246 & 0.0355 & 0.0484 & 0.0627 & 0.0847 & 0.1106 & 0.1353 & 2.8185 & 2.7353 & 2.6617 & 2.5680 & 2.4464 & 2.3008 & 2.1883 \\ \hline
\multirow{3}{*}{LPCA} &
0.06 & 0.0113 & 0.0125 & 0.0154 & 0.0168 & 0.0190 & 0.0226 & 0.0296 & 1.9785 & 1.9762 & 1.9704 & 1.9675 & 1.9632 & 1.9560 & 1.9419 \\
&0.08 & 0.0102 & 0.0115 & 0.0129 & 0.0142 & 0.0130 & 0.0120 & 0.0139 & 1.9816 & 1.9790 & 1.9763 & 1.9736 & 1.9759 & 1.9780 & 1.9743 \\
&0.10 & 0.0129 & 0.0163 & 0.0181 & 0.0215 & 0.0251 & 0.0301 & 0.0342 & 1.9778 & 1.9710 & 1.9676 & 1.9607 & 1.9533 & 1.9435 & 1.9353 \\ \hline
\multirow{3}{*}{RQMF} & 0.06 
& 0.0019 & 0.0029 & 0.0045 & 0.0064 & 0.0087 & 0.0117 & 0.0154 & 0.0269 & 0.0237 & 0.0229 & 0.0273 & 0.0296 & 0.0307 & 0.0292 \\
&0.08&  0.0023 & 0.0033 & 0.0046 & 0.0065 & 0.0090 & 0.0121 & 0.0161 & 0.0284 & 0.0273 & 0.0286 & 0.0289 & 0.0253 & 0.0242 & 0.0243 \\
&0.10&  0.0031 & 0.0040 & 0.0053 & 0.0072 & 0.0097 & 0.0127 & 0.0164 & 0.0288 & 0.0295 & 0.0308 & 0.0298 & 0.0307 & 0.0302 & 0.0325 \\
 \bottomrule
\end{tabular}}
\vspace{-4mm}
\end{table}

\begin{table}[t]
    \centering
    \caption{The comparison for elapsed-time for different methods.  \label{synthetic_time_cost}}
    \vspace{2mm}
    \resizebox{0.6\linewidth}{!}{
    \begin{tabular}{c|cccccc} \toprule
     Methods &  RSQMF  & MLS &  \zcolor{LOG-KDE} & MFIT & LPCA & RQMF \\ \hline
     Elapsed-Time  & 1.977 & 0.017 & 1.942 & 2.804 & 0.009 & 1.094 \\ \bottomrule
    \end{tabular}}
    \vspace{-4mm}
\end{table}

To evaluate the performance of various methods under different noise levels and neighborhood sizes, we conducted experiments with noise levels $\sigma$ set from $\{0.08, 0.10, 0.12\}$ and local region parameters defined as $K=4\ell+22$ for $\ell=0:6$. We set the regularization parameter $\lambda=0$ for RSQMF. Similarly, we set $\lambda=0.01$ for RQMF to mitigate the risk of overfitting. The evaluation metrics, ${\cal F}_e$ and ${\cal T}_e$, are provided in Table~\ref{synthetic}, while the computational cost, measured in \zcolor{CPU time}, is reported in Table~\ref{synthetic_time_cost}.

Our findings indicate that RSQMF consistently outperforms the other five related methods, demonstrating superior accuracy with smaller fitting errors in both ${\cal F}_e$ and ${\cal T}_e$. Notably, RSQMF exhibits heightened robustness in recovering the tangent space. This stands in contrast to \zcolor{LOG-KDE}, MFIT and LPCA, which derive their latent representations through repeated implementations of the subspace constraint mean-shift algorithm. The convergence points reached by these methods may deviate significantly from the true projection, resulting in larger errors when estimating the tangent space.  RSQMF outperforms RQMF due to its ability to independently handle and separate the tangent space from the normal space. This separation allows RSQMF to better capture local geometric structures, leading to improved performance in tasks such as manifold learning and tangent space estimation.


\subsection{Image's Variation in ${\mathbb R}^2$}
\begin{figure}[t] 
   \centering
   \includegraphics[width=0.35\linewidth]{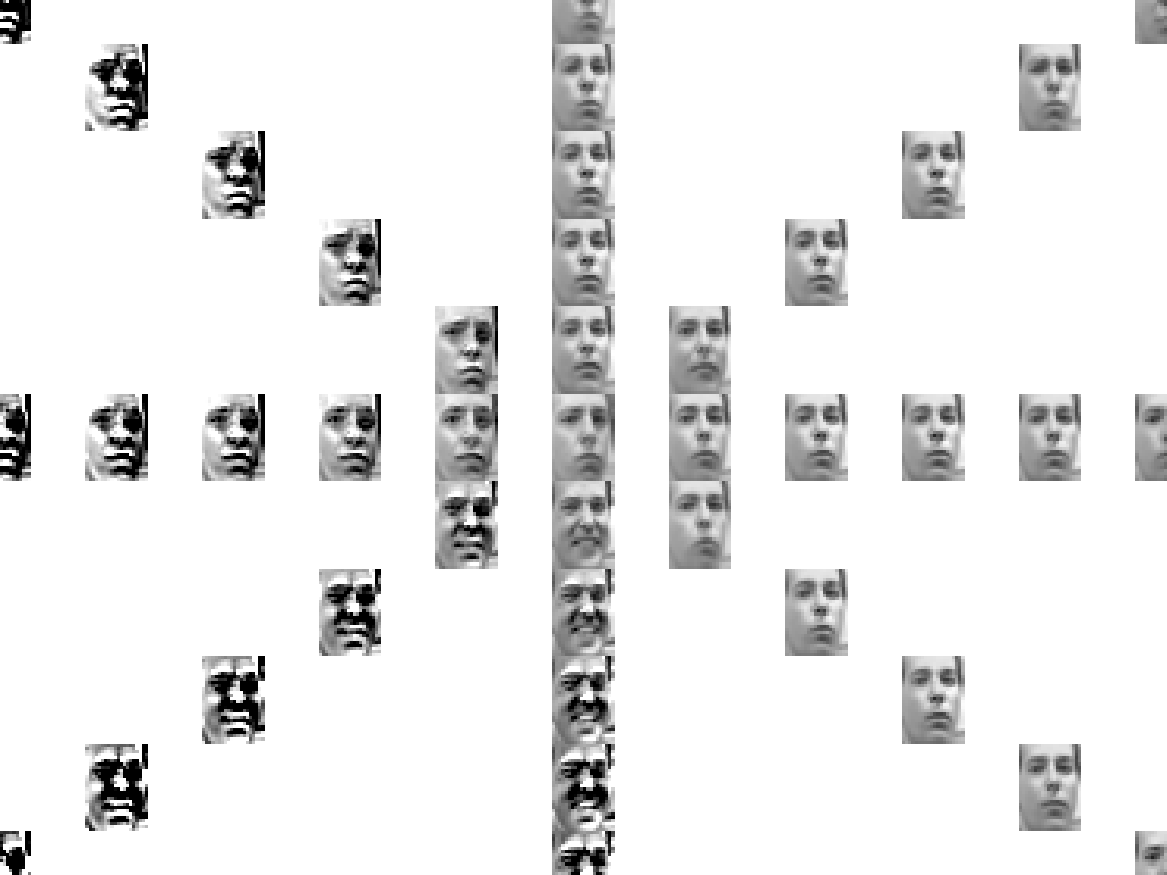} 
   \hspace{0.15\linewidth}
   \includegraphics[width=0.35\linewidth]{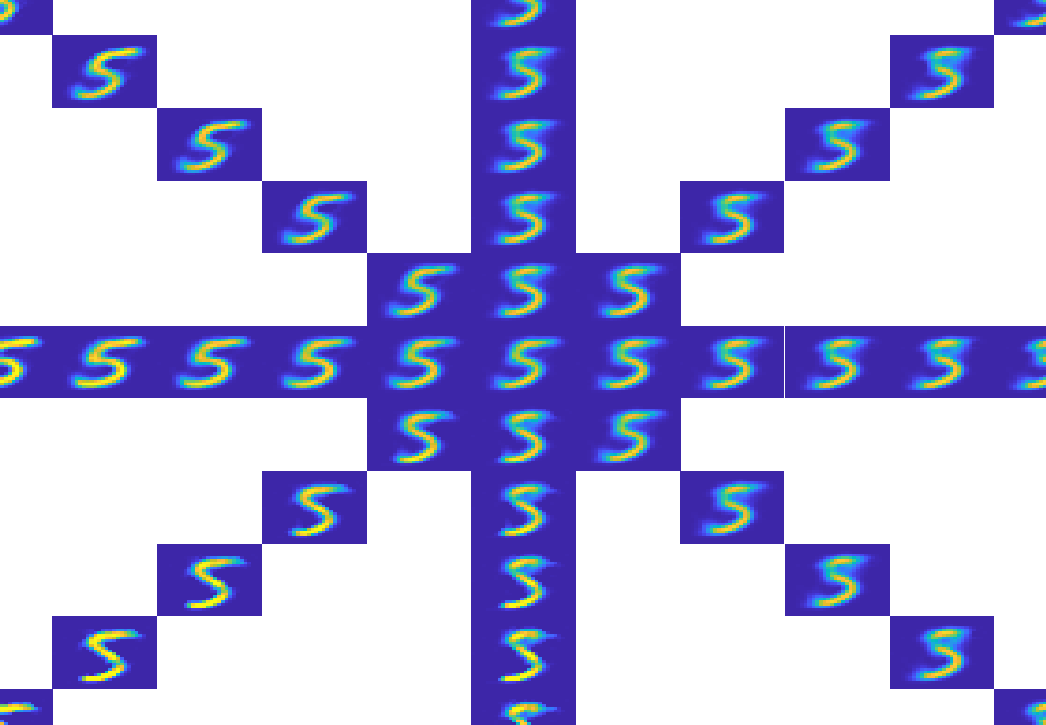} 
   \caption{The illustration of the image's variation with the change of the latent coordinate for RSQMF $f(\tau)$ on the Frey Face and Mnist dataset, where the coordinates of $\tau$ represent the location of the corresponding image. }
   \label{fig:trend}
\end{figure}
\zcolor{In this subsection}, we study the models' ability to capture the variation trend. We randomly select an image from the Frey Face dataset and utilize its 300 nearest neighbors in Euclidean space to train a quadratic factorization model $f(\tau)$ with $\tau\in {\mathbb R}^2$, $s = 3$, and $\lambda=0$. This function can be viewed as an image generator, decoding a low-dimensional vector in ${\mathbb R}^2$ into an image in ${\mathbb R}^{28\times 20}$. Next, we manually construct $\tau$ in four directions: $v_1= [1,0]$, $v_2=[0,1]$, $v_3=[1,1]$, and $v_4=[1,-1]$, by setting the local coordinate $\tau_{t,i} = \mu_i v_t$ with $\mu_i$ in the range of $[-1,1]$. Finally, we display the images of $f(\tau_{t,i})$ at the positions of $\tau_{t,i}$ in Figure \ref{fig:trend}.

As we optimize an RSQMF model with a latent dimension $d=2$, the task narrows down to interpreting the variation trends along each of the two dimensions. In the left panel of Figure \ref{fig:trend}, we observe two key patterns: First, moving from left to right along the horizontal axis reveals a gradual gender transition from a female face with glasses to a male face without glasses. Second, the direction of \zcolor{the} vertical axis clearly influences facial expressions in the Frey Face dataset\footnote{Available at https://cs.nyu.edu/~roweis/data.html}, with an increase in the vertical coordinate corresponding to a shift from a smiling expression to a more serious demeanor. In the right panel of Figure \ref{fig:trend}, we explore how the embedding coordinates in $\mathbb{R}^2$ affect font style in the MNIST dataset~\cite{mnist}. As we move along the x-axis from lower to higher values, the horizontal stroke in the handwritten digit ‘5’ transitions from long to short. Additionally, the stroke style of ‘5’ evolves from sharp to smooth as we move along the vertical axis.
\subsection{Latent Representation and Data Reconstruction\label{handwritten}}


In this subsection, we tackle a more challenging task by applying RSQMF to learn a global model on a manually selected dataset. The complexity distribution of the chosen data adds to the challenge, \zcolor{primarily due to the potential existence of the indefinite second-order fundamental form matrices in $\mathcal{A}$.}

\begin{figure}[t]
\hspace{-.11\linewidth}
\includegraphics[width=1.18\linewidth]{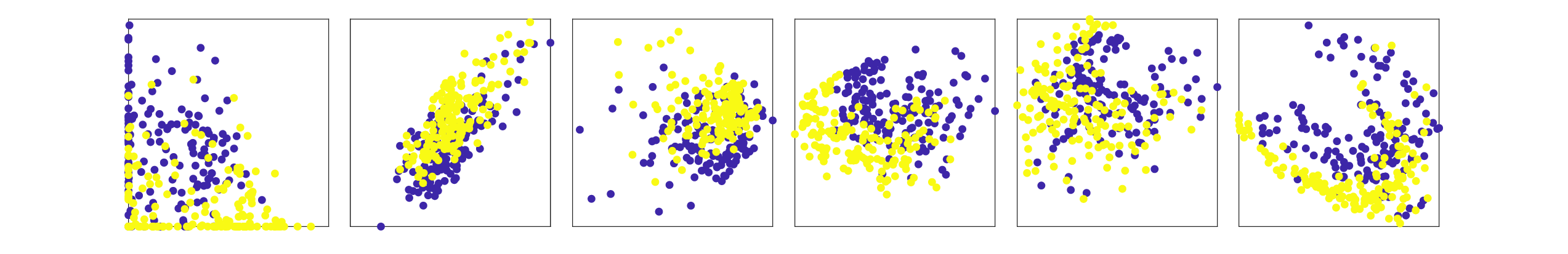}
\vspace{-12mm}
\caption{\zcolor{The comparison the embeddings for NMF, Semi-NMF, MMF, MF, QMF and RSQMF in the projected perspective of  $y-z$ and $x-z$ planes.\label{emb}}}
\end{figure}

\begin{figure}[t]
\hspace{-.03\linewidth}
\includegraphics[width=0.281\linewidth]{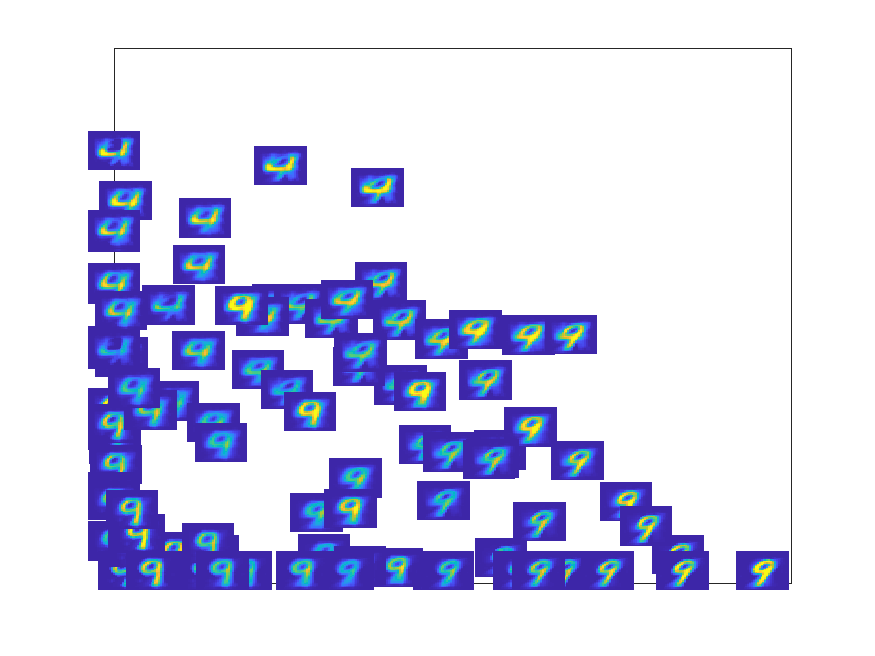}
\hspace{-.04\linewidth}
\includegraphics[width=0.281\linewidth]{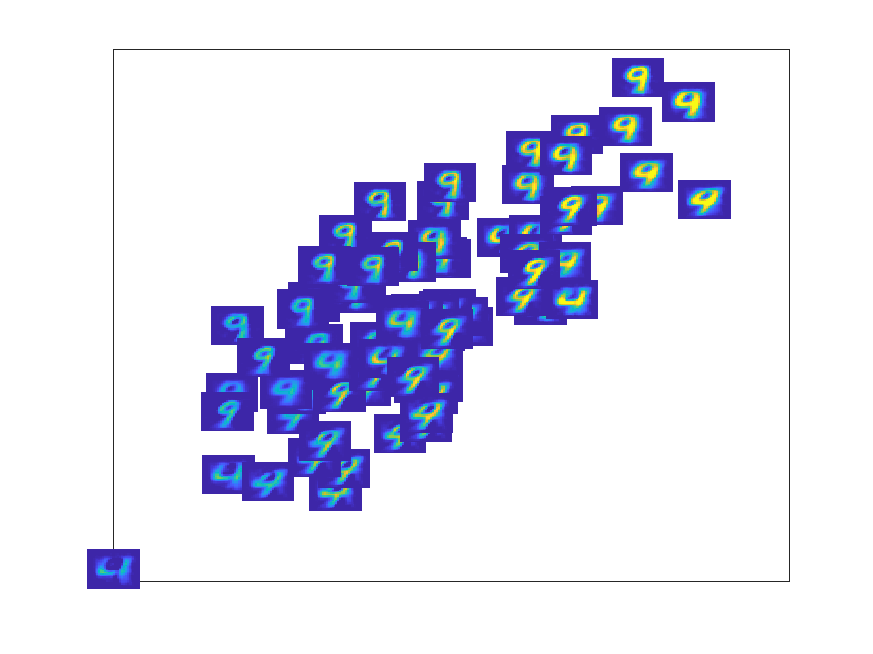}
\hspace{-.06\linewidth}
\includegraphics[width=0.281\linewidth]{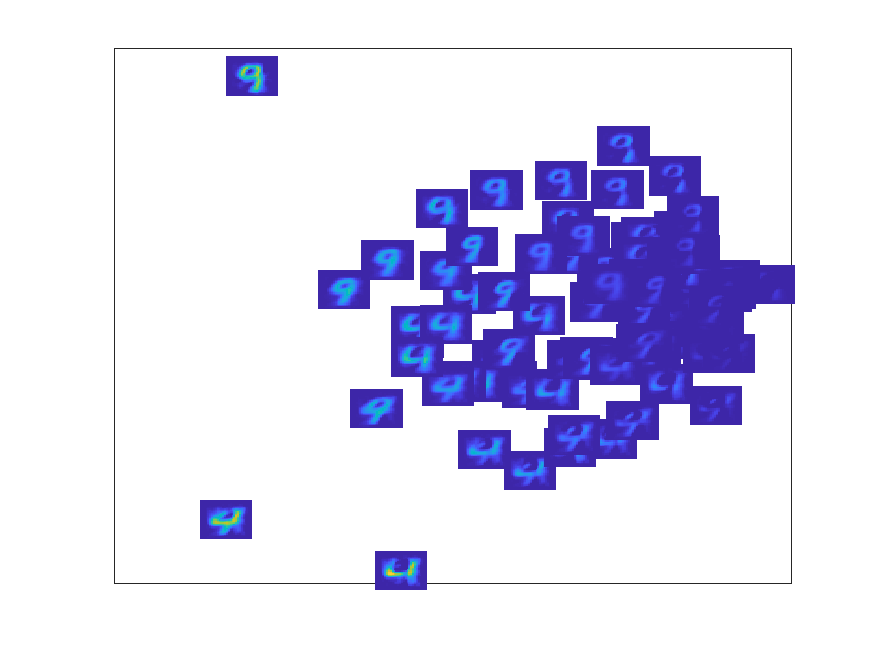}
\hspace{-.04\linewidth}
\includegraphics[width=0.281\linewidth]{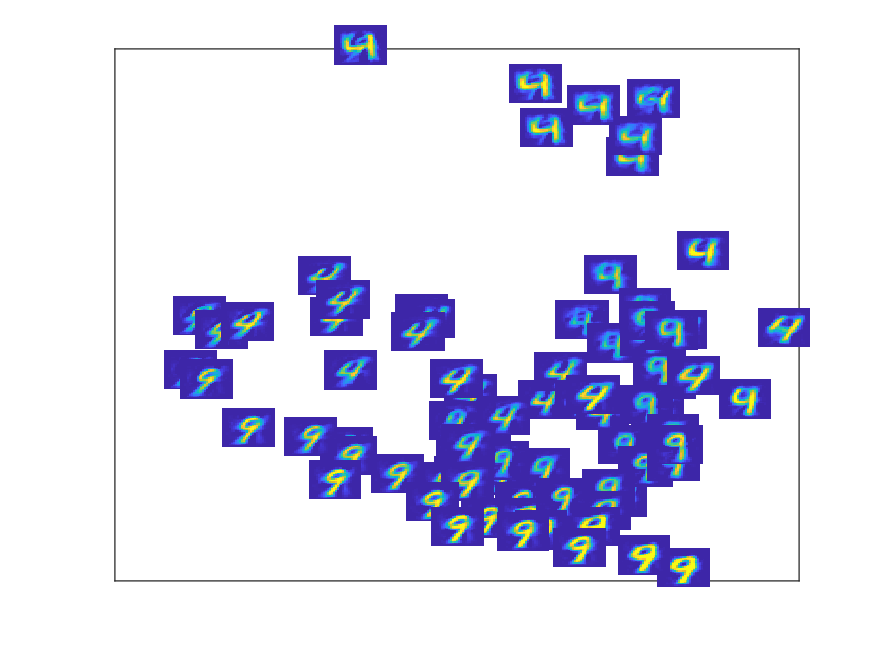}
\vspace{-12mm}
\caption{\zcolor{The reconstructed images displayed at the location of their corresponding embedding coordinates for NMF, Semi-NMF,  MMF and RSQMF.\label{reconstruction1}}}
\vspace{-4mm}
\end{figure}

To provide further detail, we assemble an image set comprising two classes, each containing 150 randomly selected images representing `4' and `9' from the MNIST~\cite{mnist} dataset. We then proceed to train RSQMF with $d=3$ on this dataset. The resulting low-dimensional embeddings of ${\tau_i}$ are visualized in Figure \ref{emb}, and the reconstructed images are depicted in Figure \ref{reconstruction1}. We
present the average fitting error measured by mean squared error and the clustering result of the embedding representation measured by accuracy (ACC) in Table \ref{error}. The elapsed CPU time for each method is also reported, with all methods evaluated under the same stopping criteria, defined as  $|\ell_{n+1} - \ell_n| \leq 10^{-5}$. \zcolor{For MMF, we set the trade-off parameter within the range $\{0.05k,k=1,...,20\}$ and constructed the connection matrix using the Gaussian kernel. The results reported correspond to the configuration achieving the highest ACC.} 
Finally, we investigate the parameter reliance and present the average fitting errors for the images corresponding to MF and RSQMF across different \zcolor{settings} of $\lambda$ and $s$ in Table \ref{impro}.

\begin{figure}[t]
\hspace{-.1\linewidth}
\includegraphics[width=1.2\linewidth]{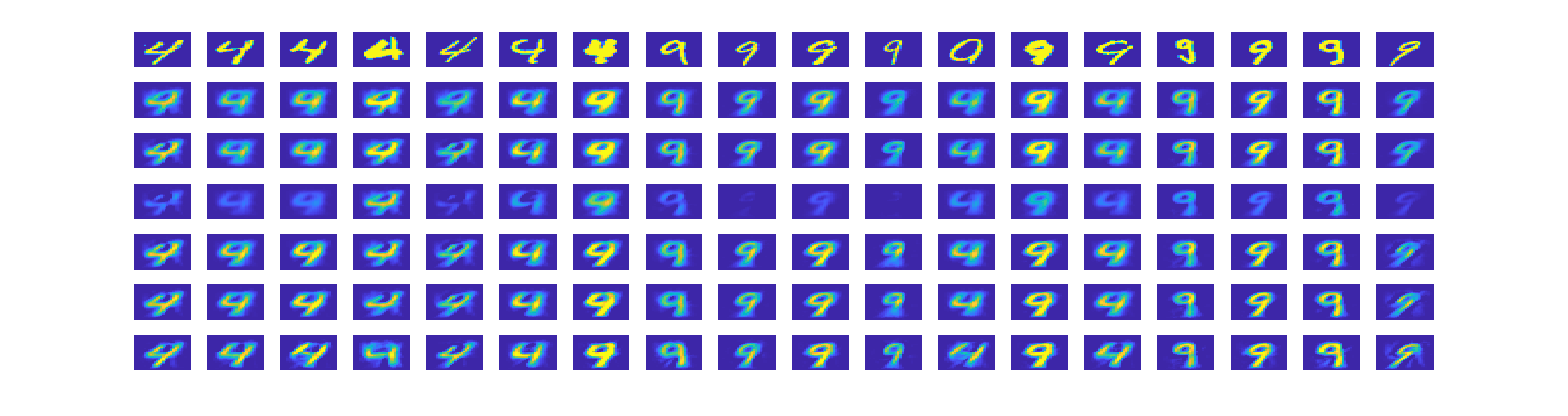}
\vspace{-14mm}
\caption{The comparison of the image reconstruction from NMF, Semi-NMF, \zcolor{MMF}, MF, RQMF and RSQMF $(\lambda=0,d=3,s=4)$ on the MNIST dataset for images corresponding to `4' and `9'.\label{reconstruction} \zcolor{The first row stands for the original images, while the second through seventh rows showcases the low-dimensional representations from NMF, Semi-NMF, \zcolor{MMF}, MF, RQMF and RSQMF, respectively.}}
\end{figure}

Upon examining Figure \ref{emb}, it is evident that the embeddings for the digits `4’ and `9’ are distinctly separated by a curve, clearly delineating the two classes with a noticeable gap \zcolor{for RSQMF}. In contrast, the embeddings generated by linear projection methods like NMF, Semi-NMF, MF, and RQMF exhibit significant overlap, making it more difficult to distinguish between the two classes. Furthermore, we illustrate the reconstructed images at their own embedding coordinates in Figure \ref{reconstruction1}, where we can observe the variation trends with the \zcolor{two-dimensional} coordinates. Finally, we compare the reconstruction capability for NMF, Semi-NMF, \zcolor{MMF}, MF, RQMF and RSQMF in Figure \ref{reconstruction}.


\begin{table}[t]
\centering
\vspace{-4mm}
\caption{The fitting error, measured by $\|f(\tau_i)-x\|_2^2$, corresponds to RSQMF for different settings of $s$ and $\lambda$, along with the relative improvement compared with MF (in brackets).\label{impro}}
\vspace{2mm}
\resizebox{\linewidth}{!}{
\begin{tabular}{c|c|c|c|c|c|c}\toprule 
$\lambda$ {\large \textbackslash}   $s$ &1&2&3&4&5&6\\ \hline
0.0& 1.787(6.2\%) & 1.700(10.1\%) & 1.643(13.1\%) & 1.606(15.1\%) & 1.818(4.5\%) & 1.773(6.6\%) \\ 
0.2& 1.812(5.3\%) & 1.727(9.8\%) & 1.677(12.4\%) & 1.643(14.1\%) & 1.625(14.9\%) & 1.713(9.4\%) \\ 
0.4& 1.823(4.8\%) & 1.751(8.8\%) & 1.714(10.6\%) & 1.672(12.9\%) & 1.659(13.6\%) & 1.650(14.1\%) \\ 
0.6& 1.831(4.5\%) & 1.768(8.0\%) & 1.736(9.6\%) & 1.698(11.6\%) & 1.688(12.1\%) & 1.677(12.8\%)  \\ 
0.8&  1.838(4.1\%) & 1.783(7.2\%) & 1.755(8.7\%) & 1.720(10.5\%) & 1.713(10.8\%) & 1.701(11.6\%)\\
\bottomrule 
\end{tabular}}
\vspace{-3mm}
\end{table}
\begin{table}[t]
\centering
\caption{The performance measured by mean squared error (MSE), accuracy (ACC) and Elapsed Time (in seconds) for \zcolor{six}  matrix factorization related methods.\label{error}}
\vspace{2mm}
\begin{tabular}{c|cccccc} \toprule
Algorithms &NMF & Semi-NMF &  \zcolor{MMF} & MF & RQMF & {RSQMF} \\ \hline
MSE &2.077 & 2.023 & \zcolor{3.103} & 1.910 &1.818 & {\bf 1.606}   \\ \hline
ACC & 0.567 & 0.765 & \zcolor{0.642} & 0.619 & 0.619 & 0.651 \\ \hline
Elapsed Time(s) & 0.13 & 0.21 & \zcolor{0.84} & 0.10 &4.75 & 1.11\\
\bottomrule
\end{tabular}
\vspace{-4mm}
\end{table}

To provide a more detailed comparison, we evaluate the performance of RSQMF by varying the parameter $\lambda$ throughout the set $\{0.2 k, \, k=0,\dots,4\}$ and the normal space parameter $s$ over the range $\{1,2,\dots,(d^2+d)/2\}$. After training RSQMF to obtain the reconstruction function $f(\tau)$, we compute the reconstruction error using the $\ell_2$ norm, defined as $\|f(\tau_i) - x_i\|_2^2$, where $f(\cdot)$ represents the reconstruction method, such as linear reconstruction by MF or orthonormal quadratic reconstruction via RSQMF. The fitting errors for RSQMF under various parameter settings are summarized in Table \ref{impro} for comparative analysis.
\zcolor{From the observations in Figure \ref{reconstruction}, Table \ref{impro} and Table \ref{error}}, we can conclude that:
\begin{itemize}[leftmargin=6mm]
\item[1.] RSQMF consistently outperforms the optimal linear model of MF and other constrained MF methods, such as \zcolor{MMF}, NMF and Semi-NMF, by achieving lower fitting errors across all parameter combinations. Additionally, as shown in Figure \ref{reconstruction}, RSQMF excels in capturing finer details compared to the competing methods, further highlighting its superior performance.

\item[2.] The fitting performances across the \zcolor{six} different methods are ranked as follows: \zcolor{$E_{\rm MMF}>E_{\rm NMF}>E_{\rm Semi-NMF}> E_{\rm MF}>E_{\rm RQMF}>E_{\rm RSQMF}$, }where $E_{\rm NMF}$ stands for the mean squared error for NMF, with similar notations used for the other methods.
\item[3.] When  $s$  is small—specifically, when  $s$  is less than 4 in this case—the performance of RSQMF tends to decline as  $\lambda$  increases. This degradation occurs because the stronger regularization term increasingly suppresses the influence of the quadratic component, leading to the model learning a flatter structure. As a result, the model’s ability to capture complex patterns diminishes, thereby reducing overall performance.
\item[4.]
As $s$ increases, for instance, when the normal space dimensional parameter $s$ takes values of 5 or 6 in this scenario, a properly selected $\lambda$ can significantly enhance the performance of RSQMF.
\end{itemize}

We provide an explanation for the observed performance ranking $\zcolor{E_{\rm MMF}}> E_{\rm NMF}>E_{\rm Semi-NMF}> E_{\rm MF}>E_{\rm RQMF}>E_{\rm RSQMF}$ in Table \ref{error} as follows: Among fixed rank matrix factorization models, unconstrained matrix factorization (MF) achieves the lowest fitting error due to its flexibility. However, when nonnegativity constraints are introduced, the models become less flexible, leading to an increase in fitting error as a trade-off. Specifically, NMF incurs a higher fitting error than Semi-NMF because it imposes an additional constraint, further limiting the model’s adaptability. 
\zcolor{Manifold regularized matrix factorization (MMF) leads to relatively large fitting errors with suboptimal clustering performance. This is primarily due to the influence of the regularization term and the difficulty of constructing a reliable connection matrix in Euclidean space using an unsupervised approach, particularly for image data.}
In contrast, the quadratic factorization model leverages nonlinear fitting in the normal space, allowing it to capture more complex data patterns and thus outperform linear factorization models. Additionally, RSQMF surpasses RQMF due to its space-separation property and the reduced number of quadratic parameters, which contribute to enhanced stability and robustness. This combination of factors allows RSQMF to achieve superior performance in fitting tasks.

From both computational cost and clustering perspectives, as shown in Table \ref{error}, we observe the following: Matrix factorization (MF) incurs the lowest computational cost, as it only requires computing a partial SVD decomposition. Both NMF and Semi-NMF are relatively faster because they involve entrywise nonnegative updating, which are computationally efficient. \zcolor{In contrast, MMF is relatively slower because it requires computing the inverse of an  $n \times n$  matrix with each update. RQMF and RSQMF also incur higher computational costs, as they involve repeated regression and projection procedures.} RSQMF demonstrates a relatively satisfactory performance in terms of ACC, though it does not perform as well as Semi-NMF. 


Finally, we provide a rationale for introducing the regularization term when $s$  is large. The parameter  $s$  represents the number of quadratic forms defined on the tangent space, and as  $s$  increases, it becomes increasingly difficult to maintain the positive definite property of  $H_{f_\lambda}(\tau)$. This challenge leads to difficulties in ensuring the convergence of iterations for solving the projection problem, often resulting in unstable solutions. By introducing the regularization parameter $\lambda $, we can more easily satisfy the conditions necessary for  $H_{f_\lambda}(\tau)$ to remain positive definite. Therefore, regularizing the model becomes essential when  $s$  is relatively large to ensure stability and reliable convergence.



\subsection{Image Refinement}
\begin{figure}[t] 
   \centering   \includegraphics[width=\linewidth]{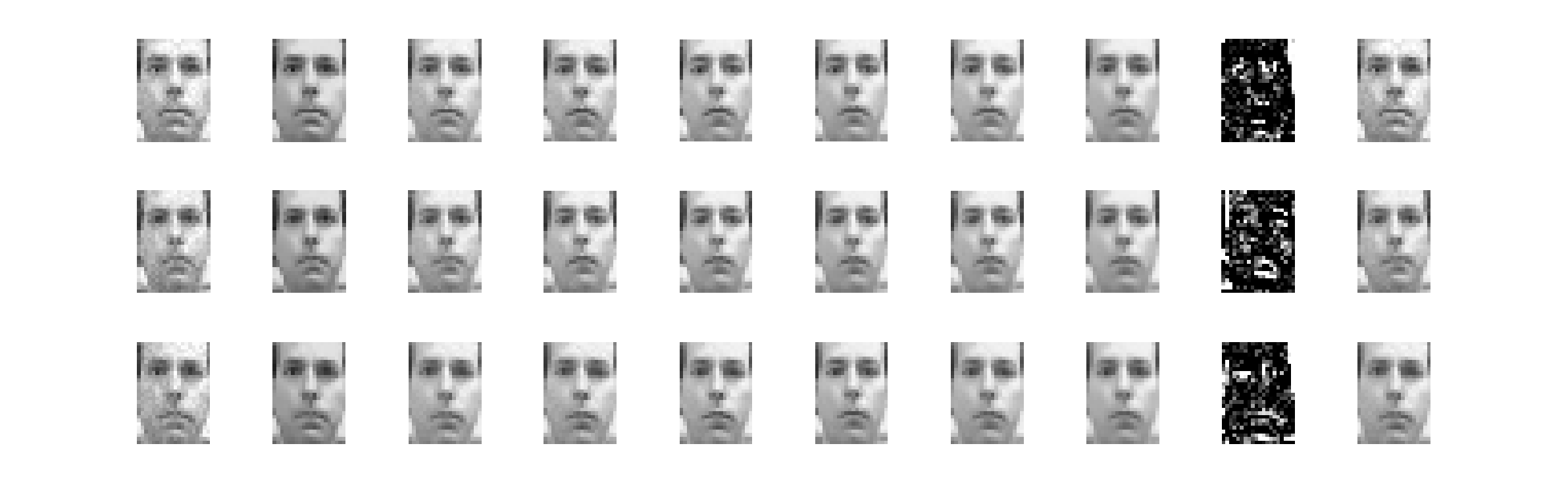} 
   \vspace{-12mm}
   \caption{The comparison of the denoising capabilities of eight different approaches on three randomly chosen images from the Frey Face dataset. The first and second columns represent the blurred and noiseless images, respectively. From the third to the tenth columns, the outputs corresponding to RSQMF, MLS, \zcolor{LOG-KDE}, MFIT, LPCA, SAME, SPCA  and RQMF are listed, respectively. }
   \label{fig:denoise_fery}
\end{figure}

In this section, we provide a comprehensive analysis of RSQMF and seven other methods when applied to denoising blurred images. To simulate the noisy input, we introduce the white noise using $I'_k = I_k + M_k$, where each ${I_k}$ and ${I'_k}$ represent images with dimensions $28 \times 20$, and $M_k$ is the noisy image. We assume each entry of $M_k$ follows a zero-mean uniform distribution, i.e., ${(M_k)}_{i,j} \sim \mathcal{U}[0,30]$. We utilize the noisy images ${I'_k}$ as input for eight denoising methods: RSQMF, \zcolor{MLS~\citep{mls}, LOG-KDE~\citep{msf}, MFIT~\citep{mfit}, LPCA~\citep{lpca}, SAME~\citep{SAME}, SPCA~\citep{Spherelets} and RQMF~\citep{Zhai}}, yielding corresponding outputs ${\hat{I}_k}$. A subset of 1000 images is randomly selected from ${I'_k}$, and their nearest $K$ neighbors are used to train an RSQMF model with $\lambda = 1$. Finally, we project the noisy images back onto the quadratic function $f(\tau)$. We explore the neighborhood parameter with various sizes $\{4k+12, k=1,...,5\}$. The resulting images from different methods are displayed in Figure \ref{fig:denoise_fery}. The mean squared error, calculated as $\frac{1}{n}\sum_k \|I'_k-\hat{I}_k\|_{\rm F}^2$, is reported in Table \ref{tab:image_denoise}. Spherelet under-performs because it heavily depends on an estimated radius, which can improperly compress or stretch the scale of the output.
Notably, RSQMF with $d=2$ and $s=3$ outperforms the related methods, achieving a smaller fitting error. This observation is coherent with our intuition that the curvature information can help us better capture the nonlinear distribution contained in the data.
\begin{table}[t]
\centering
\vspace{-2mm}
\caption{The performance comparison, measured by mean squared error (MSE), evaluates image denoising effectiveness across eight related methods under various neighborhood parameter settings.\label{tab:image_denoise}}
\vspace{2mm}
\resizebox{.85\linewidth}{!}{
\begin{tabular}{c|c|c|c|c|c|c|c|c} \toprule
{$K$}{\large \textbackslash} {Alg} & RSQMF & MLS &\zcolor{LOG-KDE} & MFIT & LPCA & SAME & SPCA& RQMF  \\ \hline
16 & \bf 1.632 & 2.475 & 2.043 & 2.561 & 2.022 & 2.112 & 342.339 & \underline{1.771} \\
20 & \bf 1.627 & 2.025 & 2.097 & 2.581 & 2.106 & 2.148 & 344.413 & \underline{1.707} \\
24 & \bf 1.652 & 2.027 & 2.301 & 2.669 & 2.111 & 2.203 & 345.367 & \underline{1.781} \\
28 & \bf 1.666 & 1.989 & 2.358 & 2.902 & 2.137 & 2.210 & 344.338 & \underline{1.717} \\
32 & \bf 1.743 & 2.044 & 2.403 & 2.894 & 2.152 & 2.276 & 339.628 & \underline{1.786} \\ \bottomrule
\end{tabular}}
\vspace{-4mm}
\end{table}
\section{Conclusion}
This paper introduces a subspace-constrained matrix factorization model, along with algorithms and theoretical insights regarding the solving process and convergence behavior. The strengths of our model can be highlighted as follows: First, the model captures nonlinear structures within the data, allowing for the extraction of richer feature details compared to linear models. Second, unlike previous quadratic matrix factorization approaches, we incorporate an additional subspace constraint that helps decouple the mutual influence between the tangent and normal spaces, improving the model’s precision. Third, by reducing the number of quadratic parameters to be optimized, our model demonstrates greater stability and is less prone to overfitting. Lastly, the curvature variables in our regularized quadratic model can be flexibly adjusted using the parameter $\lambda$, making the model highly adaptable to various tasks.

However, there are also some shortcomings and unexplored areas that warrant further investigation. First, while incorporating the quadratic form within a subspace is beneficial, we have not explored how to optimally select the dimension parameter $s$. Our experiments indicate that $s$ significantly affects the performance, and further research is needed to understand its optimal selection. Second, although the model shows strong performance across tasks, its computational process involves both outer updates for the regression and projection problem, as well as inner updates for parameters $\{\Theta, c, Q\}$ during the regression step. Due to space and time constraints, we did not delve into speed-up strategies, an area that could improve the efficiency of our approach in future work.

Several promising directions for future research emerge. First, beyond quadratic terms, more intricate structures, such as higher-order polynomials or other nonlinear functions, could be explored to better capture the underlying data distribution. Second, while our method focuses on unsupervised learning, applying it to supervised tasks, like predicting data labels, presents an exciting new direction. \zcolor{Third, the statistical properties of the SQMF model, such as those from a non-asymptotic perspective, offer potential avenues for future exploration}. Given the manifold’s complexity, there are still several unexplored aspects that could lead to novel applications and deeper insights, which we plan to investigate in future research.

\section*{Acknowledgment}
This work was supported by the National Natural Science Foundation Funds under Grants of 11801592 and 12301478.

\end{document}



\begin{frontmatter}
\title{Supplementary Material for `Subspace-Constrained Quadratic Matrix Factorization: Algorithm and Applications'}


\author[ZZ]{Zheng Zhai}
\author[HH]{Xiaohui Li}
\affiliation[ZZ]{organization={Department of Statistics, Faculty of Arts and Sciences, Beijing Normal University},
addressline={No.18 Jinfeng Road}, 
city={Zhuhai},postcode={519087}, state={Guangdong},country={China}}
\affiliation[HH]{organization={School of Mathematics and Information Sciences, Yantai University},
addressline={No.30 Laishan Qingquan Road}, 
city={Yantai},
postcode={264005}, 
state={Shandong},
country={China}}



\end{frontmatter}

\section{Proof of Lemma \ref{convex_theorem}}
\begin{proof} First, we derive several inequalities related to the spectral norm that will be utilized in our proof.
For $A_\beta^TA_\alpha$ and ${\cal A}^*({\cal A}(\alpha,\beta)-\psi_i)$, we have their spectral norms bounded by
\[
\sigma_1(A_\beta^TA_\alpha) \leq (D-d) \gamma^2 {\mathfrak b}^2, \ \ \sigma_1({\cal A}^*({\cal A}(\alpha,\beta)-\psi_i)) \leq (D-d)(\gamma^2{\mathfrak b}^2)+\|\psi_i\|_1 {\mathfrak b}.
\]
Then, we derive the Hessian of $g(\alpha,\beta)$ and demonstrate that $H_g$ is positive definite over $\cal S_A$. With a few steps' calculations, we arrive at the expression for the Hessian matrix 
\[
H_{g_i}(\alpha,\beta) = \Gamma_1(\alpha,\beta)+ \Gamma_2(\alpha,\beta).
\]
where $\Gamma_1(\alpha,\beta)$ and $\Gamma_2(\alpha,\beta)$ are defined as:
\[
\begin{aligned}
\Gamma_1(\alpha,\beta)=&\left[
    \begin{array}{cc}
       I+2A_\beta^T A_\beta  & \bf 0 \\
        \bf 0 & I+2A_\alpha^T A_\alpha 
    \end{array}
    \right], \\
    \Gamma_2(\alpha,\beta)=&\left[
    \begin{array}{cc}
       \bf 0  & 2A^T_\beta A_\alpha+2{\cal A}^*({\cal A}(\alpha,\beta)-{\psi_i}) \\
        2A^T_\alpha A_\beta+2{\cal A}^*({\cal A}(\alpha,\beta)-{\psi_i}) & \bf 0 
    \end{array}
    \right].
\end{aligned}
\]
For $\Gamma_1(\alpha,\beta)$, given that $2A_\beta^T A_\beta$ is positive semi-definite, it follows that the minimum eigenvalue satisfies $\lambda_{\min}(\Gamma_1(\alpha,\beta))\geq 1$. 
For the symmetric matrix $\Gamma_2(\alpha,\beta)$, we find that 
$
\lambda_{\max}(\Gamma_2(\alpha,\beta)) \leq 2\sigma_{\max}(A^T_\beta A_\alpha+{\cal A}^*({\cal A}(\alpha,\beta)-{\psi_i})).
$
Apply Weyl's inequality, we obtain:
\begin{equation}\label{positivedf}
\begin{aligned}
\lambda_{\min}(H_g(\alpha,\beta))\geq &\lambda_{\min}( 
\Gamma_1(\alpha,\beta))-\lambda_{\max}(\Gamma_2(\alpha,\beta))\\
=&1+2\min\{\sigma^2_{\min}(A_\beta),\sigma^2_{\min}(A_\alpha)\}-2\sigma_{\max}(A^T_\beta A_\alpha+{\cal A}^*({\cal A}(\alpha,\beta)-{\psi_i}))\\
\geq  & 1- 4(D-d)\gamma^2{\mathfrak b}^2 - 2\|{\psi_i}\|_1\mathfrak b\geq 1/2.
\end{aligned}
\end{equation}
Due to the convexity ${\cal S}_\gamma \times {\cal S}_\gamma$, the inequality in \eqref{positivedf} indicates the strong convexity of $g(\alpha,\beta)$ over ${\cal S}_\gamma \times {\cal S}_\gamma$.
\end{proof}
\section{Proof of Theorem \ref{converge}}
\begin{proof}
Suppose, to the contrary, that there exist $\alpha^* \neq \beta^*$ such that $(\alpha^*, \beta^*)$ is the optimum of $\min_{\alpha, \beta} g(\alpha, \beta)$. The symmetry of $g(\alpha, \beta)$ implies that $(\beta^*, \alpha^*)$ is another point that achieves the minimum of $g(\alpha, \beta)$. This conflicts with the notion that a convex function has a unique optimal solution on the convex set (the line $\lambda (\alpha^*, \beta^*) + (1-\lambda) (\beta^*, \alpha^*), \lambda \in [0, 1]$). Therefore, we conclude that $\alpha^* = \beta^*$. 

Since $\min_{\alpha, \beta} g(\alpha, \beta) \leq \min_\tau f(\tau)$, and we also prove that the optimal value $\alpha^*, \beta^*$ satisfies $g(\alpha^*, \beta^*) = f(\alpha^*)$, we can conclude that $\min_{\alpha, \beta} g(\alpha, \beta) = \min_\tau f(\tau)$.
\end{proof}

\section{Proof of Proposition \ref{norm}}
\begin{proof}
By setting the gradients of $g_i(\alpha, \beta_{n-1})$ and $g_i(\alpha_n, \beta)$ to zero vectors in \eqref{min_tau_n} and solving the corresponding normal equations, we have the representation of $\alpha_n$ and $\beta_n$ as follows:
\begin{equation}\label{alphabeta}
\left\{
\begin{aligned}
\alpha_n = &(2{A}_{\beta_{n-1}}{A}_{\beta_{n-1}}^T+I_d)^{-1}(2{\cal A}^*(\psi_i)\beta_{n-1}+\phi_i),\\
\beta_n = &(2{A}_{\alpha_{n}}{A}_{\alpha_{n}}^T+I_d)^{-1}(2{\cal A}^*(\psi_i)\alpha_n+\phi_i).
\end{aligned}
\right.
\end{equation}
Given that the smallest singular value of $(2{\cal A}(\beta_n){\cal A}(\beta_n)^T+I_d)$ exceeds $1$, it follows that $\|\alpha_n\|_2$ is bounded above by
\begin{equation}\label{alpha-bound}
\|\alpha_n\|_2\leq 2\|{\cal A}^*(\psi_i)\|_{\text{op}}\|\beta_{n-1}\|_2+\|\phi_i\|_2.
\end{equation}
Similarly, by applying the properties of norms and the triangle inequality to $\beta_n$ in \eqref{alphabeta}, we derive a corresponding inequality as follows:
\begin{equation}\label{beta-bound}
\|\beta_{n-1}\|_2\leq 2\|{\cal A}^*(\psi_i)
\|_{\text{op}}\|\alpha_{n-1}\|_2+\|\phi_i\|_2.
\end{equation}
By substituting \eqref{beta-bound} into \eqref{alpha-bound}, we have:
$
\|\alpha_n\|_2
\leq 4\|{\cal A}^*(\psi_i)\|_{\text{op}}^2\|\alpha_{n-1}\|_2+ 2\|{\cal A}^*(\psi_i)\|_{\text{op}}\|\phi_i\|_2+\|\phi_i\|_2.
$
which can be equivalently expressed as:
\[
\|\alpha_n\|_2+\frac{\|\phi_i\|_2}{2\|{\cal A}^*(\psi_i)\|_{\text{op}}-1} \leq (4\|{\cal A}^*(\psi_i)\|_{op}^2)(\|\alpha_{n-1}\|_2+\frac{\|\phi_i\|_2}{2\|{\cal A}^*(\psi_i)\|_{\text{op}}-1}).
\]
Thus, $\{\|\alpha_n\|_2+\frac{\|\phi_i\|_2}{2\|{\cal A}^*(\psi_i)\|_{\text{op}}-1}\}$ is monotonically decreasing with a rate equaling to $4\|{\cal A}^*(\psi_i)\|_{op}^2$ when $\|{\cal A}^*(\psi_i)\|_{\text{op}}<1$.
By applying this recursive relationship $n$ times, starting from $\alpha_0$, we obtain
\[
\|\alpha_n\|_2+\frac{\|\phi_i\|_2}{2\|{\cal A}^*(\psi_i)\|_{\text{op}}-1} \leq (\|2{\cal A}^*(\psi_i)\|_{\text{op}}^{2n})(\|\alpha_{0}\|_2+\frac{\|\phi_i\|_2}{2\|{\cal A}^*(\psi_i)\|_{\text{op}}-1}).
\]
Therefore, we can conclude that, for any $\epsilon >0$, there exists a positive integer defined as $K_0=\frac{1}{2}\log_{2\|{\cal A}^*(\psi_i)\|_{\text{op}}}\frac{2\epsilon(\|{\cal A}^*(\phi_i)\|_{\text{op}}-1)}{(2\|{\cal A}^*(\phi_i)\|_{\text{op}}-1)\|\alpha_0\|_2+\|\phi_i\|_2}$ such that, when $n>K_0$, it follows that
$\|\alpha_n\|_2\leq \frac{\|\phi_i\|_2}{1-2\|{\cal A}^*(\psi_i)\|_{\text{op}}} +\epsilon$. A similar result can be derived for $\beta_n$.
\end{proof}
\section{Proof of Theorem \ref{converge-in}}
\begin{proof}
    Firstly, we establish the convergence property for the sequence $\{\ell(Q_t,c_t,\Theta_t),t=1,2,\cdots\}$. By utilizing the algorithm outlined in Table \ref{alg:regression}, the updating sequence $\{c_{t}, Q_{t}, \Theta_{t}\}$ adheres to the following relations: 
    \begin{equation}\label{update}
    \left\{
    \begin{aligned}
        &c_{t} = \arg\min_c \ell(Q_{t-1},c,\Theta_{t-1}),\\
        & Q_{t} = \arg \min_Q \ell(Q,c_{t},\Theta_{t-1}),\\
        &\Theta_{t} = \arg\min_\Theta \ell(Q_{t},c_{t},\Theta).
    \end{aligned}
    \right.
    \end{equation}
    This implies that $\{\ell(Q_t,c_t,\Theta_t)\}$ is monotonically decreasing by:
    \begin{equation}\label{mono}
    \ell(Q_t,c_t, \Theta_t)\leq\ell(Q_t,c_t, \Theta_{t-1})\leq \ell(Q_{t-1},c_t, \Theta_{t-1})\leq \ell(Q_{t-1},c_{t-1}, \Theta_{t-1}).
    \end{equation}
    By applying the monotone convergence theorem, we can conclude that the sequence $\{\ell(Q_t, c_t, \Theta_t)\}$ converges. This conclusion arises from the fact that the loss function $\ell(Q, c, \Theta)$ is non-negative and demonstrates a decreasing property as a result of the alternating minimization strategy.
    
    From the iteration in \eqref{update}, we can conclude that the sequence $\{\Theta_{t_j}\}$ converges, as it continuously depends on $c_{t_j}$ and $Q_{t_j}$. Specifically, for any $\Phi$ such that $\lambda_{\min}(\Psi(\Phi)\Psi(\Phi)^T)>0$,
    we have $\Theta_{t_j}  = (\Psi(\Phi)\Psi(\Phi)^T)^{-1} \Psi(\Phi)R(c_{t_j})^T V_{t_j} $. Taking the limit  $j\rightarrow \infty$, we obtain $\lim_j \Theta_{t_j} = (\Psi(\Phi)\Psi(\Phi)^T)^{-1} \Psi(\Phi)R(c^*)^T V^*$.

    Secondly, we establish that $ \|c_{t_j+1}-c_{t_j}\|_2 \rightarrow 0$, $\|\Theta_{t_j+1}-\Theta_{t_j}\|_{\rm F} \rightarrow 0$ and $ \|Q_{t_j}-Q_{t_j-1}\|_{\rm F} \rightarrow 0$. Recall that $c_{t_j}$ optimizes $\min_c \ell(Q_{t_j-1},c,\Theta_{t_j-1})$ over ${\mathbb R}^D$. Then, its gradient vanishes at $c_{t_j}$. 
    From the Taylor expansion of $ \ell(Q_{t-1},c,\Theta_{t-1})$ at $c_{t_j}$ and  $\nabla_c\ell(Q_{t-1},c,\Theta_{t-1})|_{c=c_{t_j}}=\bf 0$, we have
    \[
     \ell(Q_{t_j-1},c_{t_j-1},\Theta_{t_j-1})- \ell(Q_{t_j-1},c_{t_j},\Theta_{t_j-1})= n\|c_{t_j}-c_{t_j-1}\|_2^2.
    \]
    By the inequality in \eqref{mono} and the convergence of $\{\ell(Q_t,c_t,\Theta_t)\}$, we have the difference $ \ell(Q_{t_j-1},c_{t_j-1},\Theta_{t_j-1})- \ell(Q_{t_j-1},c_{t_j},\Theta_{t_j-1}) \rightarrow 0$, which leads to
    $\|c_{t_j}-c_{t_j-1}\|_2 \rightarrow 0$ as $j\rightarrow \infty$. 
    
    Similarly, we can demonstrate that $\|\Theta_{t_j}-\Theta_{t_j-1}\|_{\rm F} \rightarrow 0$. By Taylor expansion to $\ell(Q_{t},c_{t},\Theta)$ at $\Theta_{t_j}$ and noticing that $\nabla_{\Theta} \ell(Q_{t},c_{t},\Theta)|_{\Theta=\Theta_{t_j}} = \bf 0$, we arrive at:
    \[
    \ell(Q_{t_j},c_{t_j},\Theta_{t_j-1})- \ell(Q_{t_j},c_{t_j},\Theta_{t_j})=\|(\Theta_{t_j}-\Theta_{t_j-1})\Psi(\Phi)\|_{\rm F}^2.
    \]
    From the convergence of $\{\ell(Q_t,c_t,\Theta_t)\}$ and the inequality in \eqref{mono}, we know $\|\Theta_{t_j}-\Theta_{t_j-1}\|_{\rm F} \rightarrow 0 $ as long as $\Psi(\Phi)$ has full row-rank, i.e., $\lambda_{\min}(\Psi(\Phi)\Psi(\Phi)^T)>0$.
    
    To establish that $\{Q_{t_j}\}$ also satisfies $\lim_j \|Q_{t_j} - Q_{t_j-1}\|_{\rm F} = 0$ , we can leverage the relationship of $\lim_j \Theta_{t_j} = \lim_j\Theta_{t_j-1} = \Theta^*$, $ \lim_j c_{t_j+1}= \lim_j c_{t_j} = c^*$ and the continuity property of the singular vectors. 
    Since $R(c^*)M(\Theta^*)^T$ has distinct singular values, we know that, for sufficiently large $j$, the singular vectors of $R(c_{t_j})M(\Theta_{t_j-1})^T$ and $R(c_{t_j-1})M(\Theta_{t_j-2})^T$ are both distinct. Therefore, $Q_{t_j}$ and $Q_{t_j-1}$ are uniquely determined by $R(c_{t_j})M(\Theta_{t_j-1})^T$ and  $R(c_{t_j-1})M(\Theta_{t_j-2})^T$, implying $\lim_j \|Q_{t_j}-Q_{t_j-1}\|_{\rm F} = 0$.

    Finally, we demonstrate the accumulation point $\{Q^*,t^*,\Theta^*\}$ satisfies the KKT condition $\ell(Q,c,\Theta)$. In the updating \eqref{update}, we have
    \[
    \left\{
    \begin{array}{l}
         \nabla_c \ell(Q_{t_j-1},c,\Theta_{t_j-1})|_{c_{t_j}} = 0,\\
          Q_{t_j}= U \Lambda V^T,\\
          Q_{t_j}^TQ_{t_j} = I_{d+s},\\
         \nabla_{\Theta} \ell(Q_{t_j},c_j,\Theta)|_{\Theta_{t_j}} = 0.
    \end{array}
    \right.
    \]
    Here, $\Lambda$ is the dual variable and $U, V$ are the unitary matrices such that $U \Lambda V^T = R(c_{t_j})M(\Theta_{t_j-1})^T $. Letting $j\rightarrow\infty$ and applying the result of $\lim_j \|Q_{t_j-1}-Q_{t_j}\|_{\rm F}= 0, \lim_j \|c_{t_j-1}-c_{t_j}\|_2= 0,\lim_j \|\Theta_{t_j-1}-\Theta_{t_j}\|_{\rm F}= 0$ , we arrive at our conclusion.
\end{proof}
\section{Proof of Theorem \ref{converge-out}}
\begin{proof}
Our proof consists of the following four steps:

First, we prove $\ell(\Theta,c,Q,\Phi)$ is strongly convex for $\Phi$ in the product set of  ${\cal B}_1\times {\cal B}_2\times...\times {\cal B}_n$.
Utilizing the block form of the Hessian matrix $\nabla\Phi^2(\ell(\Theta,c,Q,\Phi))$ and the Hessian matrix $H_{f_i}(\tau)$ in \eqref{gH}, we find that:
\[
\begin{aligned}
\lambda_{\min}(\nabla_\Phi^2(\ell(\Theta,c,Q,\Phi))) =& \min_i \lambda_{\min}(H_{f_i}(\tau))\\
\geq & 2 + 4\lambda_{\min}(2A_\tau^T A_\tau+{\cal A}^*({\cal A}(\tau,\tau)-\psi_i)))\\
\geq & 2-4\sigma_{\max}((2A_\tau^T A_\tau+{\cal A}^*({\cal A}(\tau,\tau)-\psi_i)))))\geq 1.
\end{aligned}
\]
Therefore, $\ell(\Theta,c,Q,\Phi)$ is strongly convex over the set of  ${\cal B}_1\times {\cal B}_2\times...\times {\cal B}_n$.

Second, we prove that $\lim_k \Phi_{j_k} = \Phi^*$. By the continuity of the Hessian matrix function, for sufficiently large $k$, $\ell(\Theta_{j_k},c_{j_k},Q_{j_k},\Phi)$ is strongly convex. Newton's iteration converges to the unique solution for the problem $\min_\Phi\ell(\Theta_{j_k},c_{j_k},Q_{j_k},\Phi)$ due to the strong convexity of $\ell(\Theta_{j_k},c_{j_k},Q_{j_k},\Phi)$. Letting $k\rightarrow \infty$, we have $\lim_k \Phi_{j_k}$ converges to $\Phi^* = \arg\min_\Phi \ell(\Theta^*,c^*,Q^*,\Phi)$.

Third, we prove $\lim_k \Phi_{j_k}-\Phi_{j_k-1}=0$. By the optimality condition $\nabla_\Phi \ell(\Theta_{j_k},c_{j_k},Q_{j_k},\Phi)=\bf 0$ and the strong convexity condition, we have:
\[
\begin{aligned}
\ell(\Theta_{t_k},c_{t_k},Q_{t_k},\Phi_{t_k-1}) \geq &\ell(\Theta_{t_k},c_{t_k},Q_{t_k},\Phi_{t_k}) + \frac{1}{2} vec(\Phi_{t_k}-\Phi_{t_k-1}) H_{\Phi_{t_k}} vec(\Phi_{t_k}-\Phi_{t_k-1})\\
\geq &\ell(\Theta_{t_k},c_{t_k},Q_{t_k},\Phi_{t_k})+\frac{1}{2}\lambda_{\min}(H_{\phi_{t_k}})\|\Phi_{t_k}-\Phi_{t_k-1}\|_{\rm F}^2\\
\geq
& \ell(\Theta_{t_k},c_{t_k},Q_{t_k},\Phi_{t_k})+\frac{1}{2}\|\Phi_{t_k}-\Phi_{t_k-1}\|_{\rm F}^2.
\end{aligned}
\]
Therefore,
$\|\Phi_{t_k}-\Phi_{t_k-1}\|_{\rm F}^2 \leq 2 (\ell(\Theta_{t_k},c_{t_k},Q_{t_k},\Phi_{t_k})- \ell(\Theta_{t_k},c_{t_k},Q_{t_k},\Phi_{t_k}))\rightarrow 0$. As a result, $\lim_k \Phi_{t_k} = \lim_k \Phi_{t_k-1} = \Phi^*$.

Finally, using the convergence condition $\lim_k \Phi_{t_k-1} = \lim_k \Phi_{t_k} = \Phi^*$ together with the
optimal condition of 
\[
\begin{aligned}
(\Theta_{t_k},c_{t_k},Q_{t_k}) &= \arg\min_{\Theta_{t_k},c_{t_k},Q_{t_k}^TQ_{t_k} = I_D} \ell(\Theta,c,Q,\Phi_{t_k-1}),\\
\Phi_{t_k} &= \arg\min_\Phi \ell(\Theta_{t_k},c_{t_k},Q_{t_k},\Phi).
\end{aligned}
\]
We  conclude that accumulation point satisfies the KKT condition of the minimization problem $\min_{\Theta,c,Q,\Phi}\ell(\Theta,c,Q,\Phi)$. 
\end{proof}